\documentclass[sigconf]{acmart}
\AtBeginDocument{%
  }
\DeclareUnicodeCharacter{2212}{-}

\usepackage{graphicx}
\usepackage{subcaption}
\usepackage{multirow}
\usepackage{tcolorbox}
\usepackage{makecell}
\usepackage{booktabs}
\usepackage{multirow}

\usepackage{siunitx}
\setcopyright{acmlicensed}
\copyrightyear{2018}
\acmYear{2018}
\acmDOI{XXXXXXX.XXXXXXX}
\acmConference[Conference acronym 'XX]{Make sure to enter the correct
  conference title from your rights confirmation email}{June 03--05,
  2018}{Woodstock, NY}
\acmISBN{978-1-4503-XXXX-X/2018/06}

\begin{document}

\title[Are LLMs Ready for TOON?]{Are LLMs Ready for TOON? Benchmarking Structural Correctness–Sustainability Trade-offs in Novel Structured Output Formats}

\author{Elio Masciari}
\email{elio.masciari@unina.it}
\affiliation{%
  \institution{University of Naples Federico II}
  \city{Naples}
  \country{Italy}
}

\author{Vincenzo Moscato}
\email{vincenzo.moscato@unina.it}
\affiliation{%
  \institution{University of Naples Federico II}
  \city{Naples}
  \country{Italy}}

\author{Enea Vincenzo Napolitano}
\email{eneavincenzo.napolitano@unina.it}
\affiliation{%
  \institution{University of Naples Federico II}
  \city{Naples}
  \country{Italy}}

\author{Gian Marco Orlando}
\email{gianmarco.orlando@unina.it}
\affiliation{%
  \institution{University of Naples Federico II}
  \city{Naples}
  \country{Italy}}

\author{Marco Perillo}
\email{marc.perillo@studenti.unina.it}
\affiliation{%
  \institution{University of Naples Federico II}
  \city{Naples}
  \country{Italy}}

\author{Diego Russo}
\email{diego.russo@unibg.it}
\affiliation{%
  \institution{University of Bergamo}
  \city{Dalmine (BG)}
  \country{Italy}}

\renewcommand{\shortauthors}{Masciari et al.}

\begin{abstract}
  
\end{abstract}



\received{20 February 2007}
\received[revised]{12 March 2009}
\received[accepted]{5 June 2009}

\begin{abstract}
Large Language Models (LLMs) are increasingly required to generate structured, machine-readable outputs for downstream systems. While recent benchmarks have focused on evaluating the structural correctness of such outputs, the environmental impact of inference for different output formats has largely been overlooked. In this paper, we argue that structured output formats should be assessed not only in terms of correctness, but also with respect to their environmental efficiency. To this end, we introduce a sustainability-aware evaluation framework for structured generation that measures token usage, generation time, and estimated carbon emissions. Within this framework, we propose the Environment-Aware Generation Correctness Score (GCS$_{\text{env}}$), a unified metric that integrates structural correctness with carbon-aware efficiency. Using this framework, we systematically benchmark the novel TOON format against established representations (JSON, XML, YAML) across multiple LLMs spanning different architectures and parameter scales.

Our results reveal a consistent trade-off: TOON yields markedly more compact outputs and lower emissions, but lower structural correctness when models lack native support. We show that increased model capacity reduces this gap and that environment-aware scoring can shift format rankings depending on deployment priorities. highlighting the need for sustainability-inclusive benchmarking and provides empirical evidence that compact representations such as TOON can offer practical advantages in large-scale, carbon-conscious LLM deployments.
\end{abstract}

\maketitle

\section{Introduction}

Large Language Models (LLMs) are increasingly deployed within automated pipelines, developer tools, and machine-to-machine workflows, where they are expected to produce structured outputs that are directly consumed by downstream components. In these settings, model quality is no longer determined solely by the semantic plausibility of generated natural language, but also by the ability to reliably produce syntactically valid and structurally compliant outputs. Even minor violations of structural constraints can propagate through dependent components, resulting in parsing failures, execution errors, or costly regeneration steps.

This shift has made structured output generation a central evaluation problem, motivating the development of benchmarks and metrics that explicitly assess format compliance and structural correctness. However, existing evaluation methodologies predominantly focus on correctness-related dimensions, implicitly assuming output formats to be neutral with respect to computational cost. In parallel, the widespread deployment of LLM-based systems has amplified concerns about their environmental impact, leading to the development of the Green AI paradigm \cite{barbierato2024toward, liu2024green}. While the carbon footprint of model training has been extensively studied \cite{faiz2023llmcarbon, xu2024research}, the costs associated with inference remain comparatively under-explored despite being the dominant source of emissions in real-world applications \cite{jegham2025hungry, ding2024sustainable}. This is a particularly critical issue for structured generation tasks. Different output formats may encode identical semantic content while requiring substantially different numbers of generated tokens, decoding times, and amounts of energy. Ignoring such representational differences risks overlooking a key factor that directly affects the sustainability of LLM-based systems at scale.

In this work, we argue that structured output formats should be evaluated not only in terms of structural correctness, but also with respect to their environmental efficiency. From this perspective, compact and token-efficient representations are not merely an optimization detail, but a critical means for reducing inference-time emissions. We investigate this hypothesis by focusing on TOON \cite{lafalcetoon}, a novel structured representation designed to provide more compact encodings than widely adopted formats such as JSON, XML, and YAML. Specifically, we examine whether TOON can deliver tangible sustainability benefits when used as an LLM output format, and under which conditions current LLMs can reliably generate it.

To this end, we introduce a sustainability-aware evaluation framework for structured output generation. Based on existing correctness-oriented benchmarks \cite{StructEval}, our framework augments traditional evaluation with explicit measurements of token usage, generation time, and estimated carbon emissions, computed exclusively during the decoding phase. This design choice enables us to isolate the environmental cost of the output representation itself, independently of prompt verbosity or instruction complexity. This allows us to make fair comparisons between formats that encode equivalent semantic content, but differ in representational compactness. Within this framework, we propose a novel metric, the \emph{Environment-Aware Generation Correctness Score} ($GCS_{\text{env}}$), which integrates structural correctness with carbon-aware efficiency into a single, interpretable score. By normalizing emissions with respect to a reference emission factor, $GCS_{\text{env}}$ remains comparable across hardware configurations and experimental settings, supporting transparent and reproducible evaluation. Using the proposed framework, we conduct a systematic benchmark comparison of JSON, XML, and YAML against TOON across a diverse set of contemporary LLMs spanning multiple architectures and parameter scales. This allows us to jointly analyze the impact of output format choice and model capacity on structural reliability, efficiency, and environmental sustainability.

Our analysis is guided by three research questions: \textit{(i)} how the choice of output format affects structural correctness and environmental impact; \textit{(ii)} to what extent increasing model capacity mitigates the lack of native support for novel formats such as TOON; and \textit{(iii)} how sensitive format comparisons are to the relative importance assigned to sustainability through environment-aware evaluation.

Overall, this work provides a principled assessment of the trade-offs between structural reliability and environmental efficiency in structured output generation, highlighting the conditions under which novel compact representations such as TOON can be considered practical alternatives to established strctured formats and underscoring the importance of sustainability-aware evaluation in realistic deployment scenarios.

\subsubsection*{\textbf{Contributions of this work}}

In summary, this paper makes the following contributions:

\begin{itemize}
    \item We introduce one of the first sustainability-aware benchmarking frameworks for structured output generation, explicitly integrating inference-time efficiency and environmental impact into format evaluation.
    \item We propose the \emph{Environment-Aware Generation Correctness Score} ($GCS_{\text{env}}$), a unified metric that enables explicit and interpretable trade-offs between structural correctness and environmental efficiency.
    \item We present a large-scale empirical evaluation of TOON as a target structured output format, comparing it against JSON, XML, and YAML across multiple LLM families and parameter scales.
\end{itemize}


\section{Background and Related Work}

The generation of structured outputs has become increasingly important in modern natural language processing (NLP) systems, especially as LLMs are now integrated into software pipelines, data processing tools, and machine-to-machine communication interfaces. Understanding the characteristics of widely used data formats, the benchmarks that evaluate structured output production, and the emerging role of sustainability considerations is therefore essential to contextualize the need for a new format such as TOON and for a sustainability-oriented evaluation methodology.

\subsection{Structured Data Formats and Applications}

Structured data formats like JSON, XML, and YAML have become fundamental for representing and exchanging data in a machine-readable way. JSON is a lightweight key-value based format, widely adopted in web APIs and data interchange due to its simplicity and native support in many programming languages \cite{RFC8259}. JSON is prevalent in NLP pipelines for tasks such as information extraction and question answering, where models output structured results that downstream systems can easily parse. Its syntax is rigid, which ensures interoperability but also means that any minor deviation by an LLM (e.g., a missing comma or quote) can render the output invalid. XML, in contrast, uses hierarchical tags and attributes to represent nested structures \cite{W3CXML}. XML saw extensive use in earlier NLP systems and supports schemas (DTD/XSD) to enforce correctness. However, XML’s verbose syntax and strict closing-tag requirements can be challenging for LLMs to adhere to exactly, and its verbosity led to JSON overtaking it for most web and NLP uses.

YAML is a human-friendly format often used for configuration files and data serialization \cite{YAML2009}. It allows indentation-based nesting and is more flexible about quoting and data types than JSON. This human-readability makes YAML attractive for developers, and some NLP frameworks use YAML for model or experiment configuration. But for generated outputs, YAML’s flexibility can introduce ambiguity or inconsistent interpretation, which is problematic for automated parsing. 

These structural and syntactic differences strongly influence token production, error rates, and reparsing costs in LLM outputs, reinforcing the need for both a more token-efficient format such as TOON and a dedicated sustainability-driven metric capable of comparing formats in terms of their environmental footprint.

\subsection{Existing Benchmarks for Structured Output Evaluation}

Early work on evaluating LLM outputs mostly focused on semantic accuracy and natural language quality, with little emphasis on format correctness. However, recently researchers have introduced benchmarks to specifically assess the syntactic and semantic correctness of structured outputs from language models. For example, IFEval was proposed to test how well LLMs adhere to various instruction constraints, including output format requirements \cite{Zhou2023IFEval}. Similarly, InFoBench targets complex instructions and includes a subset of tests for format compliance, introducing metrics like a "decomposed requirements following ratio" to quantify adherence to output specifications \cite{Qin2024InfoBench}. 

Recognizing the need for dedicated evaluation of structured outputs, SoEval was introduced by Liu et al. (2024) as one of the first benchmarks centered on structured output generation \cite{Liu2024SoEval}. It revealed that while leading LLMs can produce simple JSON or XML for straightforward tasks, they struggle with deeper nested structures. JSONSchemaBench takes a rigorous approach by leveraging thousands of real-world JSON schemas as test problems \cite{Geng2025JSONSchemaBench}. It evaluates LLMs on their ability to generate outputs that are not only syntactically valid JSON but also satisfy complex schema constraints.

The StructEval benchmark is another recent comprehensive effort: it evaluates an LLM’s capability to produce a broad spectrum of structured formats \cite{Yang2025StructEval}. StructEval defines tasks in two categories: generation tasks, where the model must create a structured output from a natural language specification, and conversion tasks, where the model translates one structured format to another, covering over 18 formats. 

FOFO extends evaluation to domain-specific formats like legal documents and logs \cite{Xia2024FOFO}. These benchmarks consistently find that even advanced models may output malformed structured data, and methods like constrained decoding or self-verification are needed to improve format compliance.

\subsection{Motivations for a Sustainability-Driven Format Comparison}

The rise of large-scale models has raised concerns about their energy consumption and carbon footprint, giving rise to the field of Green AI, which calls for AI research that considers energy efficiency and environmental impact as primary metrics alongside performance \cite{Schwartz2020}. Pioneering works quantified the resource requirements of NLP models: Strubell et al. (2019) estimated that training a single big Transformer model can emit CO$_2$ equivalent to several automobiles' lifetime emissions \cite{Strubell2019}. Schwartz et al. emphasized the importance of efficiency and reporting of compute cost (FLOPs, energy, CO$_2$), while Henderson et al. (2020) proposed guidelines for systematic reporting \cite{Henderson2020}.

Tools such as experiment impact trackers and emissions calculators have emerged to help researchers measure kilowatt-hours and translate them into CO$_2$e (carbon dioxide equivalent). Industry studies confirmed that efficient training strategies, greener data centers, and optimized model architectures can significantly reduce emissions. Patterson et al. reported that choosing a carbon-efficient data center and time of day for training can drastically cut emissions \cite{Patterson2021}. 

Inference, not just training, is energy-intensive, particularly in LLMs production deployments. Hence, optimizing output format generation may impact total energy use. Certain formats might be easier for LLMs to generate correctly and efficiently, requiring fewer tokens or corrections. Comparing formats through a sustainability lens can reveal tradeoffs in computational and environmental cost. Our study embraces this perspective, aligning format evaluation with sustainability metrics such as energy use and emissions, thus expanding Green AI principles to structured output generation.

\section{Methodology}

\subsection{Overview}

Our methodology builds upon StructEval \cite{StructEval}, a structured evaluation framework designed to assess LLMs by systematically probing their capabilities across well-defined task objectives and output constraints. \cite{StructEval} systematically evaluates structural fidelity across diverse output formats through two paradigms: \textit{i)} generation tasks, producing structured output from natural language prompts, and \textit{ii)} conversion tasks, translating between structured formats.

In this work, we adopt the generation task defined in \cite{StructEval}, where models are prompted to produce outputs that strictly conform to a predefined structured schema. In this setting, the objective is to evaluate the model’s ability to faithfully translate natural language instructions into machine-readable structured representations, preserving both syntactic validity and semantic alignment with the input specification. Following \cite{StructEval}, we consider generation as a controlled task with explicit format constraints, which allows evaluation to be grounded on objective criteria rather than subjective judgments.

Within this scope, we restrict our analysis to widely adopted serialization formats. Specifically, we employed JSON, XML, and YAML, which are explicitly supported in \cite{StructEval} and are commonly used as target representations in data interchange, configuration, and tool-calling scenarios. To these formats, we add TOON\footnote{\url{https://toonformat.dev/}} as a novel structured output representation. TOON is designed to express the same underlying information content while enabling more compact, token-efficient, and structurally explicit encodings. This makes it particularly suitable for studying generation efficiency under fixed semantic constraints, without altering the task objective itself. Importantly, TOON is treated as an additional target format within the same generation task, rather than as a new task or capability, ensuring full comparability with existing StructEval-based benchmarks.

\subsection{Metrics}

To comprehensively evaluate structured generation across different output formats, we adopt a set of complementary metrics capturing efficiency, sustainability, and structural correctness. All metrics are computed at the level of a single generation instance and are aggregated across the evaluation set. Importantly, in contrast to end-to-end measurements that conflate prompt processing with decoding, all metrics considered in this study are computed exclusively over the LLM output generation phase, measured from the emission of the first token to the completion of the final token. This methodological choice allows us to isolate the contribution of the output representation itself, independently of prompt-related overhead.

This distinction is particularly relevant in our experimental setting. While widely adopted formats such as JSON, XML, and YAML are implicitly supported by the training data of contemporary LLMs and typically require minimal or no explicit formatting instructions, the TOON format is not natively known by the evaluated models and therefore necessitates more detailed prompt-level specifications of its structural constraints. In our experiments, the instructions required for correct TOON formatting are provided directly within the prompt and are derived from the official TOON format specification\footnote{\url{https://toonformat.dev/guide/format-overview.html}}. Including prompt processing in metric computation would thus introduce a systematic bias against TOON, as observed differences would partially reflect instructional verbosity rather than intrinsic properties of the generated representation. By restricting all measurements to the decoding phase, we ensure a fair and controlled comparison across formats that encode equivalent semantic content but differ in structural compactness and tokenization behavior.

In the following paragraphs, we detail the metrics adopted in this study, describing how each captures a specific aspect of structured generation quality, efficiency, and environmental impact.

\paragraph{\textbf{Duration}}
We measure the generation duration as the wall-clock time elapsed from the moment the model starts producing the output until the generation is completed. This metric captures latency introduced by the output format itself, independently of upstream prompt processing, and provides a direct indicator of responsiveness under identical task conditions.

\paragraph{\textbf{Number of Generated Tokens ($N_T$)}}
We record the total number of tokens $N_T$ generated by the LLM for each output. This metric represents the primary, format-agnostic indicator of computational effort during decoding and plays a central role in our efficiency analysis. Since all structured formats considered in this study encode the same semantic content, differences in $N_T$ directly reflect representational compactness and tokenization behavior. 

\paragraph{\textbf{Estimated Carbon Emissions (CE)}}
We quantify sustainability in terms of the estimated carbon emissions (in kgCO\textsubscript{2}e) produced during the output generation phase, denoted as $CE$. In our setting, carbon emissions are derived from the number of generated tokens $N_T$, by translating the token-level computational cost incurred during decoding into energy consumption and subsequently into carbon emissions using hardware and location emission factors, following established practices in Green AI research \cite{Henderson2020}. Consistent with prior work on inference-time efficiency, we compute $CE$ exclusively over the decoding phase, as inference has been shown to represent a significant and recurring source of energy consumption in production-scale LLM deployments \cite{samsi2023words}. By grounding emission estimates in token counts, this formulation ensures that observed differences in $CE$ directly reflect the impact of output format choice rather than prompt-related overhead.

\paragraph{\textbf{Render Score}}
Following \cite{StructEval}, we compute the Render Score, a binary metric (0 or 1) which assesses whether the generated output can be successfully parsed and rendered according to the target structured format specification, indicating whether the generated code can be successfully loaded or rendered without syntax errors. This metric captures hard format compliance, penalizing malformed outputs that fail to instantiate a valid structured object, and is independent of semantic correctness.

\paragraph{\textbf{Syntax Score}}
We additionally report the Syntax Score defined in \cite{StructEval}, which evaluates the syntactic validity of the generated output with respect to the formal grammar of the target format. Unlike the Render Score, which reflects end-to-end parsability, the Syntax Score focuses on local structural correctness (e.g., existence of required keys, valid key–value structures), providing a finer-grained signal of format adherence. This metric is calculated as the percentage of key-value pairs satisfied by the generated output format.

\subsection{Evaluation Criteria}

Our evaluation criteria integrate structural correctness, as defined by \cite{StructEval}, with an environmental efficiency component introduced in this work, resulting in a unified score that jointly captures output validity and sustainability. We denote the original metric as the Generation Correctness Score (GCS), and its environment-aware extension as $GCS_{env}$, which incorporates carbon-aware efficiency into generation correctness.

\paragraph{\textbf{Generation Correctness Score (GCS)}}
Following \cite{StructEval}, the Generation Correctness Score ($GCS$) reflects the overall structural correctness of the generated output. Given a generated output, the $GCS$ is measured as:

\begin{equation}
    \text{GCS} = \alpha \cdot \text{Render Score} + \beta \cdot \text{Syntax Score}
\end{equation}

\paragraph{\textbf{Environmental Efficiency Score (EES)}}
To account for the environmental impact of structured generation, we introduce an environmental efficiency score, denoted as Environmental Efficiency Score ($EES$), which quantifies the carbon intensity of the generation process. We define the the amount of CO\textsubscript{2} emitted to generate 1000 output tokens as:

\begin{equation}
    X = \frac{1000 \cdot CE{}}{N_T}
\end{equation}

We normalize this quantity using a reference emission factor $X_{ref}$, defined as the estimated CO\textsubscript{2} emissions per 1000 tokens produced by a state-of-the-art LLM. Using this reference, we define the environmental efficiency score as:

\begin{equation}
    EES = \frac{1}{1 + \frac{X}{X_{\text{ref}}}}
\end{equation}

This formulation yields a bounded score in the interval [0,1], where values closer to 1 indicate lower carbon intensity relative to the reference model. Importantly, this definition does not rely on dataset-specific normalization or task-dependent baselines, ensuring comparability across formats and experimental settings.

\paragraph{\textbf{Environment-Aware Generation Correctness Score ($GCS_{env}$).}}
We combine structural correctness and environmental efficiency into a single environment-aware final score, denoted as $GCS_{env}$. This score is defined as:

\begin{equation}
    GCS_{env} = (1 - \gamma) \cdot GCS + \gamma \cdot EES
\end{equation}

where $\gamma \in [0,1]$ controls the relative importance of sustainability with respect to structural correctness. By construction, $GCS_{env}$ preserves the original $GCS$ evaluation signal while explicitly incorporating the carbon intensity of output generation. This enables a principled comparison of structured output formats that encode equivalent semantic content but differ in representational compactness, tokenization behavior, and environmental impact.

\section{Experiments}

\subsection{Experimental Setup}

\paragraph{\textbf{Hardware Configuration.}}
All experiments were conducted on a system with a 10th Gen Intel  i9-10980XE processor, 256 GB of RAM, and an NVIDIA RTX A6000  with 48 GB memory.

\paragraph{\textbf{Model Selection.}}
We conduct our experiments on a diverse set of LLMs in order to assess the robustness of our findings across heterogeneous generation behaviors. Specifically, we evaluate the following models: GPT-oss-20B and GPT-oss-120B \cite{agarwal2025gpt}, Gemma 3 models at 4B, 12B, and 27B parameters \cite{team2025gemma}, Mistral 7B \cite{jiang2023mistral7b}, Llama 3.3 70B\footnote{\url{https://www.llama.com/docs/model-cards-and-prompt-formats/llama3_3/}}, and Qwen 3 4B \cite{yang2025qwen3}.

This selection is motivated by three main considerations. First, it enables a systematic analysis across model scales, ranging from compact models (4B parameters) to large-scale models exceeding 70B parameters, allowing us to study how structured generation quality and environmental efficiency vary as a function of model dimensions. Second, the chosen models belong to different model families, each characterized by distinct architectural choices, training corpora, and instruction-tuning strategies. This diversity mitigates the risk that our results are tied to idiosyncratic behaviors of a single model lineage. Third, the selected models represent a mix of instruction-tuned and general-purpose open-weight models, reflecting realistic deployment scenarios in which structured output generation is performed under varying degrees of alignment and formatting reliability. Notably, Qwen 3 4B is included as it emerges as the best-performing open-source model in the StructEval benchmark \cite{StructEval} in terms of structured generation correctness.

By evaluating this heterogeneous model set, we aim to ensure that the observed trade-offs between structural correctness, efficiency, and environmental impact are not artifacts of a specific model or scale, but rather reflect general properties of structured generation under contemporary LLMs.

\paragraph{\textbf{Parameter Settings.}}
Carbon emissions $E_{tot}$ associated with output generation were estimated using CodeCarbon\footnote{\url{https://github.com/mlco2/codecarbon}}. Consistent with our evaluation protocol, emissions were measured exclusively during the output generation phase, starting from the first generated token and ending with the final token.

The $GCS$ was computed by preserving the weighting parameters of the original StructEval benchmark \cite{StructEval}, with weights set to $\alpha = 0.2$ and $\beta = 0.8$. Accordingly, $GCS$ is defined as:

\begin{equation}
    \text{GCS} = 0.2 \cdot \text{Render Score} + 0.8 \cdot \text{Syntax Score}
\end{equation}

For the environmental efficiency component, we set the reference emission factor to $X_{ref} = 0.001719 \; kgCO_2e$ which corresponds to the emission estimate for GPT-4 per 1000 generated output tokens reported by the LLM Carbon Calculator\footnote{\url{https://llmemissions.com/}}. This value is derived from empirical measurements of token-level energy consumption reported by \cite{samsi2023words} on different LLMs, and adjusted to reflect the improved energy efficiency of modern GPU hardware. Finally, the $GCS_{env}$ was computed by combining correctness and environmental efficiency with a weighting factor $\gamma$. In all experiments, we set $\gamma = 0.5$, thus assigning equal importance to structural correctness and carbon-aware efficiency. Therefore, $GCS_{env}$ is defined as:

\begin{equation}
    GCS_{env} = 0.5 \cdot GCS + 0.5 \cdot EES
\end{equation}

\paragraph{\textbf{Evaluation Protocol.}}
We consider a fixed set of 50 generation instances for each format comparison. For every instance, the same natural language query is issued once for each original structured format (JSON, XML, YAML) and once for its corresponding TOON representation. This paired evaluation design ensures a controlled comparison: each TOON output is directly matched with its format-equivalent baseline, allowing observed differences in correctness, token usage, generation time, and environmental impact to be attributed solely to the output representation rather than to task variability. All reported results are obtained by aggregating metrics over these paired generations.


\subsection{Benchmark Overview and Research Questions} \label{sec:results_overview}

This study aims to systematically analyze the trade-offs between structural correctness and environmental efficiency in structured output generation with large language models. In particular, we investigate \textit{(i)} the impact of output format choice on correctness and sustainability, \textit{(ii)} the role of model capacity in supporting novel structured representations, and \textit{(iii)} how environment-aware evaluation criteria influence the comparative assessment of output formats. To this end, we structure our analysis around the following research questions (RQs):

\begin{itemize}
    \item \textbf{RQ1 (Format-level impact):}  
    How does adopting a novel structured output format such as TOON affect structural correctness and environmental impact compared to established formats (JSON, XML, YAML)?

    \item \textbf{RQ2 (Role of model capacity):}  
    To what extent can increased model capacity compensate for the absence of native support for novel structured output formats such as TOON?

    \item \textbf{RQ3 (Sensitivity to environment-aware evaluation):}  
    How does varying the importance assigned to environmental efficiency affect the comparative evaluation of structured output formats?
\end{itemize}

\subsection{Aggregate Results Across Models} \label{sec:results_aggregate}

To address \textbf{RQ1}, we begin by presenting aggregate benchmark results obtained by summarizing evaluation metrics across all considered models. This analysis aims to capture global, model-agnostic trends that are stable across different model families and parameter scales, thereby providing a high-level characterization of the impact of output format on structured generation. Importantly, this aggregation is intended to highlight consistent patterns rather than to replace model-level analyses, which are examined in detail in the subsequent sections.

Table \ref{tab:aggregate_all_models} reports the aggregate comparison between TOON and each baseline structured format (JSON, XML, and YAML) across all evaluation metrics. For each metric, we report the mean and standard deviation computed across models.

\begin{table*}[t]
    \centering
    \resizebox{0.7\textwidth}{!}{%
    \begin{tabular}{l|cc|cc|cc}
    \hline
    \textbf{Metric} 
    & \textbf{JSON} & \textbf{TOON} 
    & \textbf{XML}  & \textbf{TOON} 
    & \textbf{YAML} & \textbf{TOON} \\
    \hline

    Duration $\downarrow$
    & \makecell{\small\textbf{9.724}\\[-3pt]{\scriptsize(± 2.675)}}
    & \makecell{\small 10.007\\[-3pt]{\scriptsize(± 3.58)}}
    & \makecell{\small 13.84\\[-3pt]{\scriptsize(± 4.397)}}
    & \makecell{\small\textbf{10.878}\\[-3pt]{\scriptsize(± 4.332)}}
    & \makecell{\small\textbf{9.363}\\[-3pt]{\scriptsize(± 2.711)}}
    & \makecell{\small 10.909\\[-3pt]{\scriptsize(± 3.683)}} \\[5pt]

    CE ($\times 10^{-4}$) $\downarrow$
    & \makecell{\small 0.422\\[-3pt]{\scriptsize(± 0.306)}}
    & \makecell{\small\textbf{0.072}\\[-3pt]{\scriptsize(± 0.045)}}
    & \makecell{\small 0.672\\[-3pt]{\scriptsize(± 0.428)}}
    & \makecell{\small\textbf{0.083}\\[-3pt]{\scriptsize(± 0.047)}}
    & \makecell{\small 0.381\\[-3pt]{\scriptsize(± 0.244)}}
    & \makecell{\small\textbf{0.088}\\[-3pt]{\scriptsize(± 0.062)}} \\[5pt]

    $N_T$ $\downarrow$
    & \makecell{\small 296.15\\[-3pt]{\scriptsize(± 27.428)}}
    & \makecell{\small\textbf{217.94}\\[-3pt]{\scriptsize(± 33.826)}}
    & \makecell{\small 472.528\\[-3pt]{\scriptsize(± 76.65)}}
    & \makecell{\small\textbf{239.205}\\[-3pt]{\scriptsize(± 51.286)}}
    & \makecell{\small 275.518\\[-3pt]{\scriptsize(± 38.991)}}
    & \makecell{\small\textbf{233.5}\\[-3pt]{\scriptsize(± 37.034)}} \\[5pt]

    Render Score $\uparrow$
    & \makecell{\small\textbf{0.990}\\[-3pt]{\scriptsize(± 0.021)}}
    & \makecell{\small 0.630\\[-3pt]{\scriptsize(± 0.322)}}
    & \makecell{\small\textbf{0.985}\\[-3pt]{\scriptsize(± 0.021)}}
    & \makecell{\small 0.622\\[-3pt]{\scriptsize(± 0.312)}}
    & \makecell{\small\textbf{0.985}\\[-3pt]{\scriptsize(± 0.03)}}
    & \makecell{\small 0.635\\[-3pt]{\scriptsize(± 0.319)}} \\[5pt]

    Syntax Score $\uparrow$
    & \makecell{\small\textbf{0.802}\\[-3pt]{\scriptsize(± 0.06)}}
    & \makecell{\small 0.484\\[-3pt]{\scriptsize(± 0.284)}}
    & \makecell{\small\textbf{0.457}\\[-3pt]{\scriptsize(± 0.057)}}
    & \makecell{\small 0.33\\[-3pt]{\scriptsize(± 0.171)}}
    & \makecell{\small\textbf{0.845}\\[-3pt]{\scriptsize(± 0.041)}}
    & \makecell{\small 0.471\\[-3pt]{\scriptsize(± 0.287)}} \\[5pt]

    GCS $\uparrow$
    & \makecell{\small\textbf{0.840}\\[-3pt]{\scriptsize(± 0.05)}}
    & \makecell{\small 0.513\\[-3pt]{\scriptsize(± 0.291)}}
    & \makecell{\small\textbf{0.563}\\[-3pt]{\scriptsize(± 0.047)}}
    & \makecell{\small 0.389\\[-3pt]{\scriptsize(± 0.198)}}
    & \makecell{\small\textbf{0.872}\\[-3pt]{\scriptsize(± 0.036)}}
    & \makecell{\small 0.504\\[-3pt]{\scriptsize(± 0.291)}} \\[5pt]

    EES $\uparrow$
    & \makecell{\small 0.926\\[-3pt]{\scriptsize(± 0.052)}}
    & \makecell{\small\textbf{0.980}\\[-3pt]{\scriptsize(± 0.015)}}
    & \makecell{\small 0.928\\[-3pt]{\scriptsize(± 0.041)}}
    & \makecell{\small\textbf{0.979}\\[-3pt]{\scriptsize(± 0.016)}}
    & \makecell{\small 0.928\\[-3pt]{\scriptsize(± 0.041)}}
    & \makecell{\small\textbf{0.978}\\[-3pt]{\scriptsize(± 0.018)}} \\[5pt]

    $GCS_{env}$ $\uparrow$
    & \makecell{\small\textbf{0.883}\\[-3pt]{\scriptsize(± 0.025)}}
    & \makecell{\small 0.747\\[-3pt]{\scriptsize(± 0.139)}}
    & \makecell{\small\textbf{0.745}\\[-3pt]{\scriptsize(± 0.026)}}
    & \makecell{\small 0.684\\[-3pt]{\scriptsize(± 0.095)}}
    & \makecell{\small\textbf{0.901}\\[-3pt]{\scriptsize(± 0.024)}}
    & \makecell{\small 0.741\\[-3pt]{\scriptsize(± 0.139)}} \\

    \hline
    \end{tabular}}
    \caption{Aggregate paired comparison between TOON and standard structured formats across all evaluated models. Mean shown above, standard deviation in parentheses below.}
    \label{tab:aggregate_all_models}
\end{table*}

\begin{figure*}[t]
    \centering
    \begin{subfigure}[t]{0.32\textwidth}
        \centering
        \includegraphics[width=\linewidth]{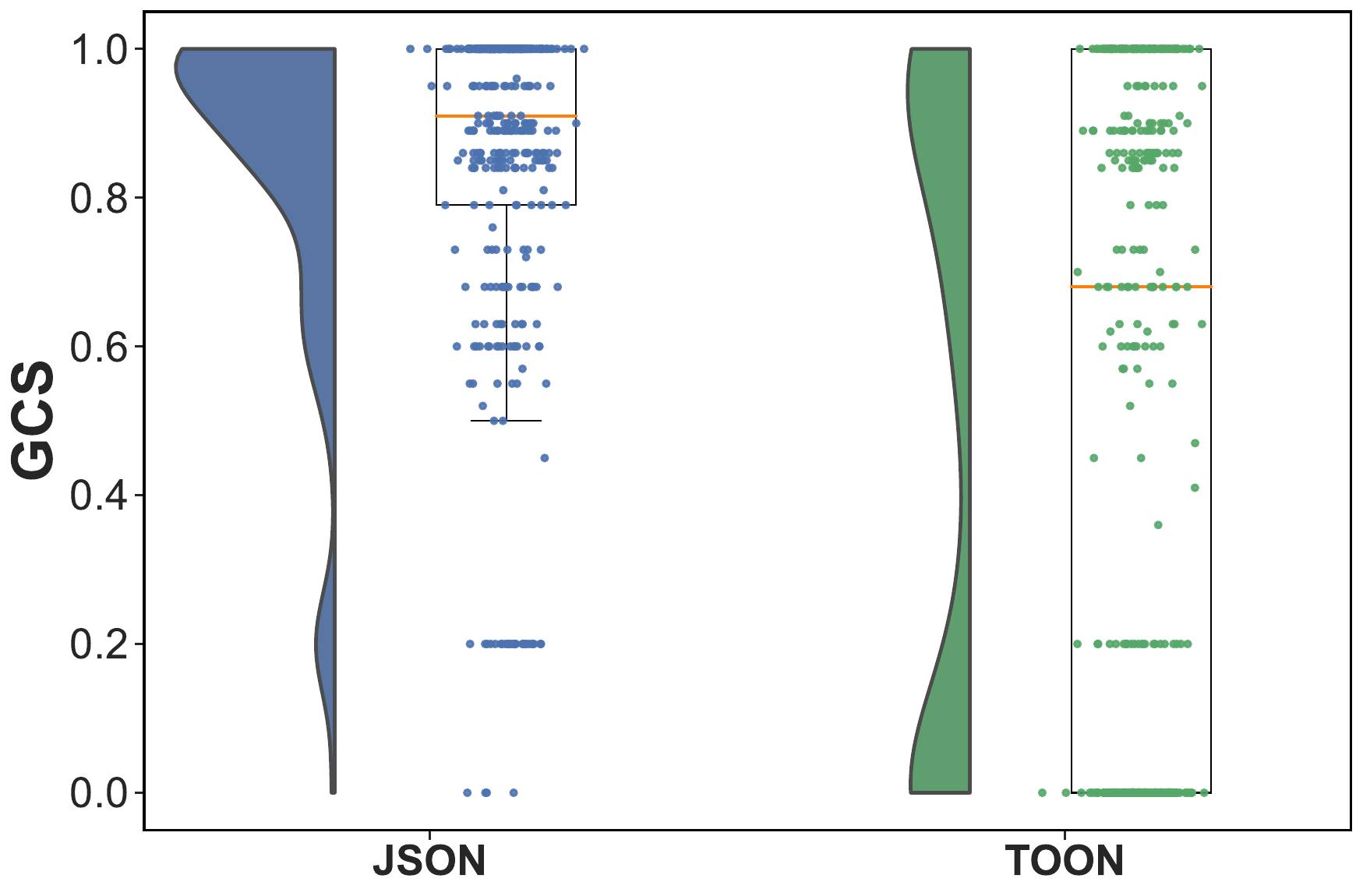}
        \caption{Structural correctness comparison (GCS): JSON vs TOON.}
        \label{fig:img11_GCS}
    \end{subfigure}\hfill
    \begin{subfigure}[t]{0.32\textwidth}
        \centering
        \includegraphics[width=\linewidth]{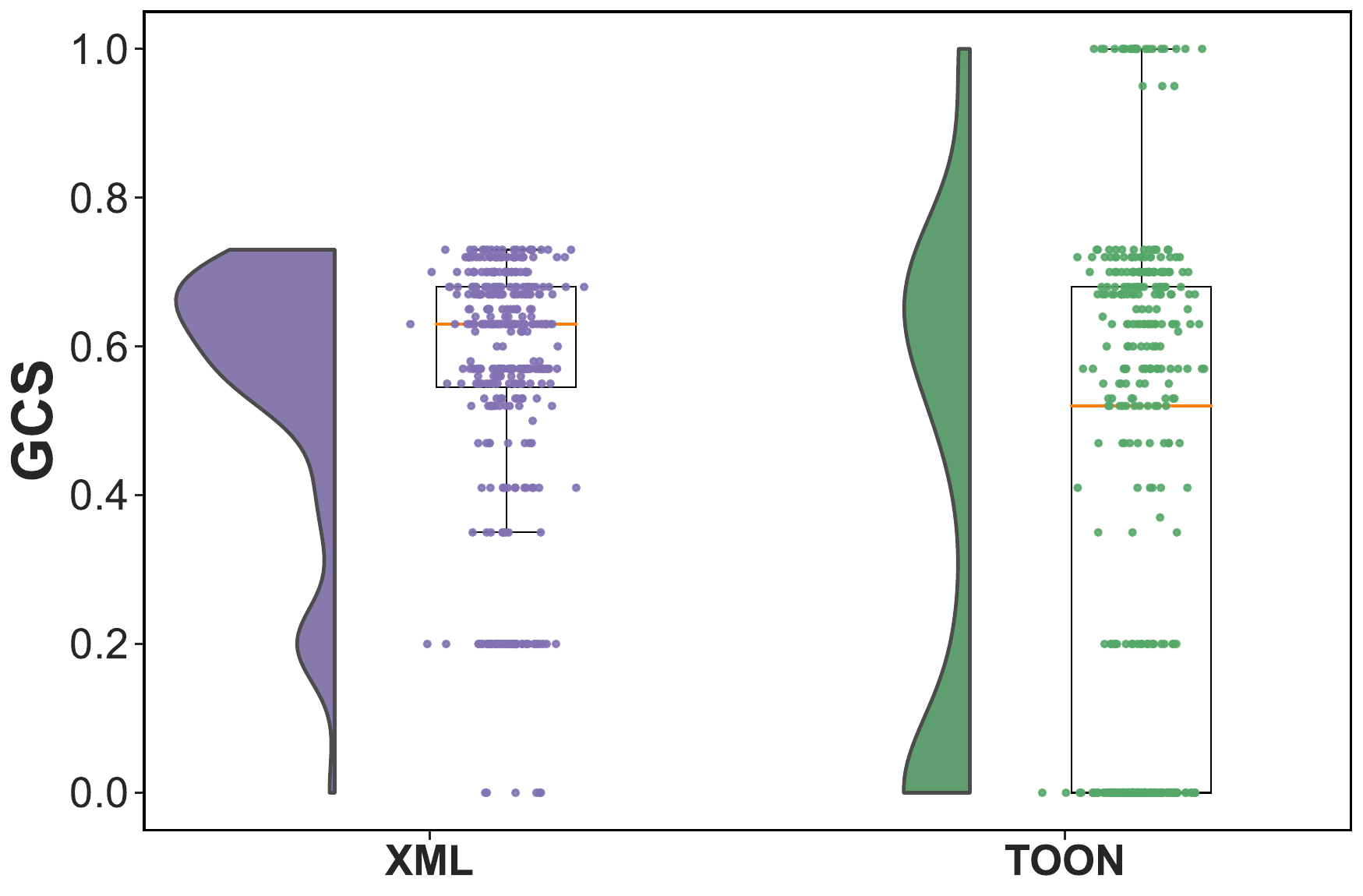}
        \caption{Structural correctness comparison (GCS): XML vs TOON.}
        \label{fig:img12_GCS}
    \end{subfigure}\hfill
    \begin{subfigure}[t]{0.32\textwidth}
        \centering
        \includegraphics[width=\linewidth]{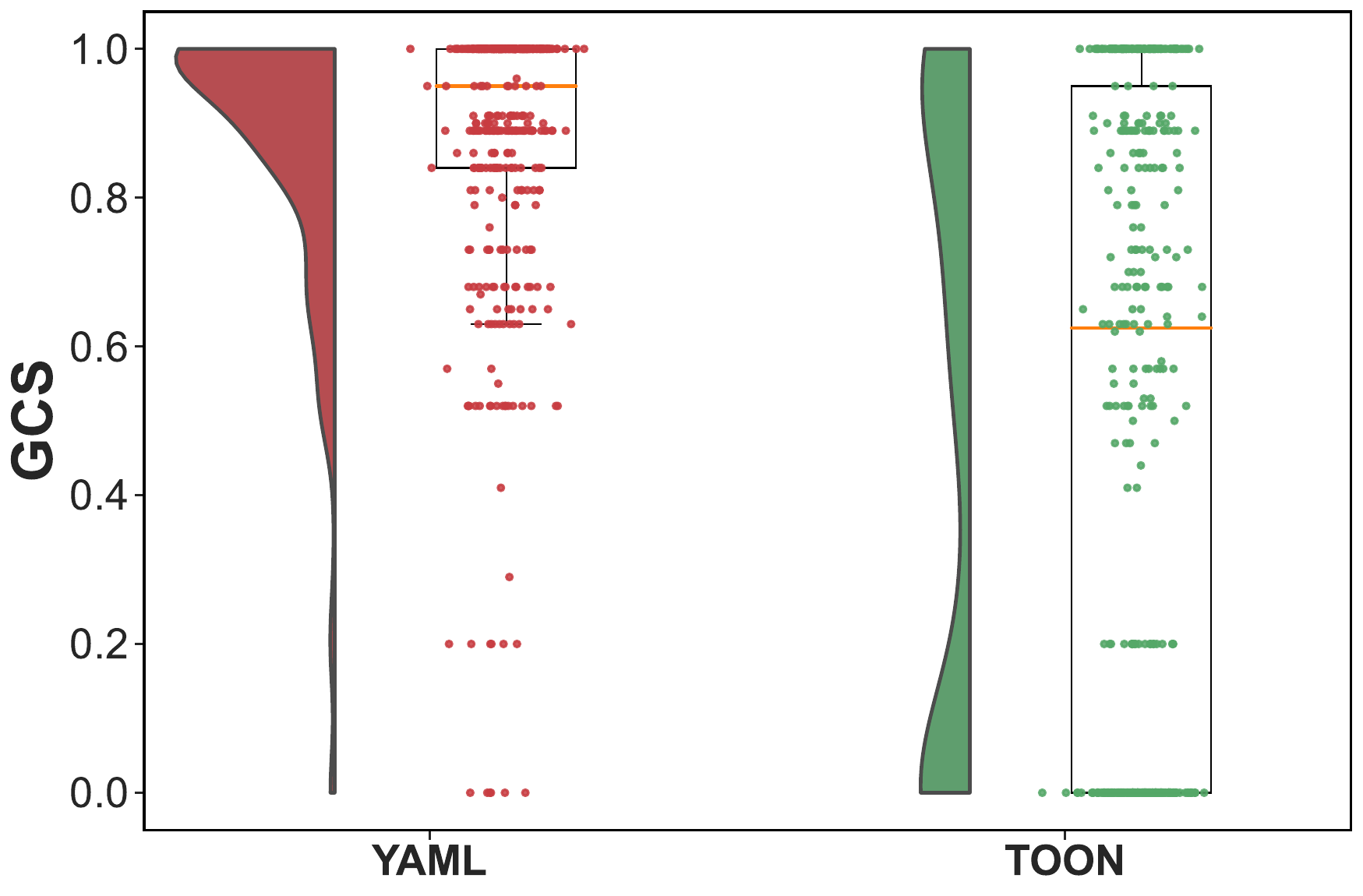}
        \caption{Structural correctness comparison (GCS): YAML vs TOON.}
        \label{fig:img13_GCS}
    \end{subfigure}
    \vspace{0.8em}
    \begin{subfigure}[t]{0.32\textwidth}
        \centering
        \includegraphics[width=\linewidth]{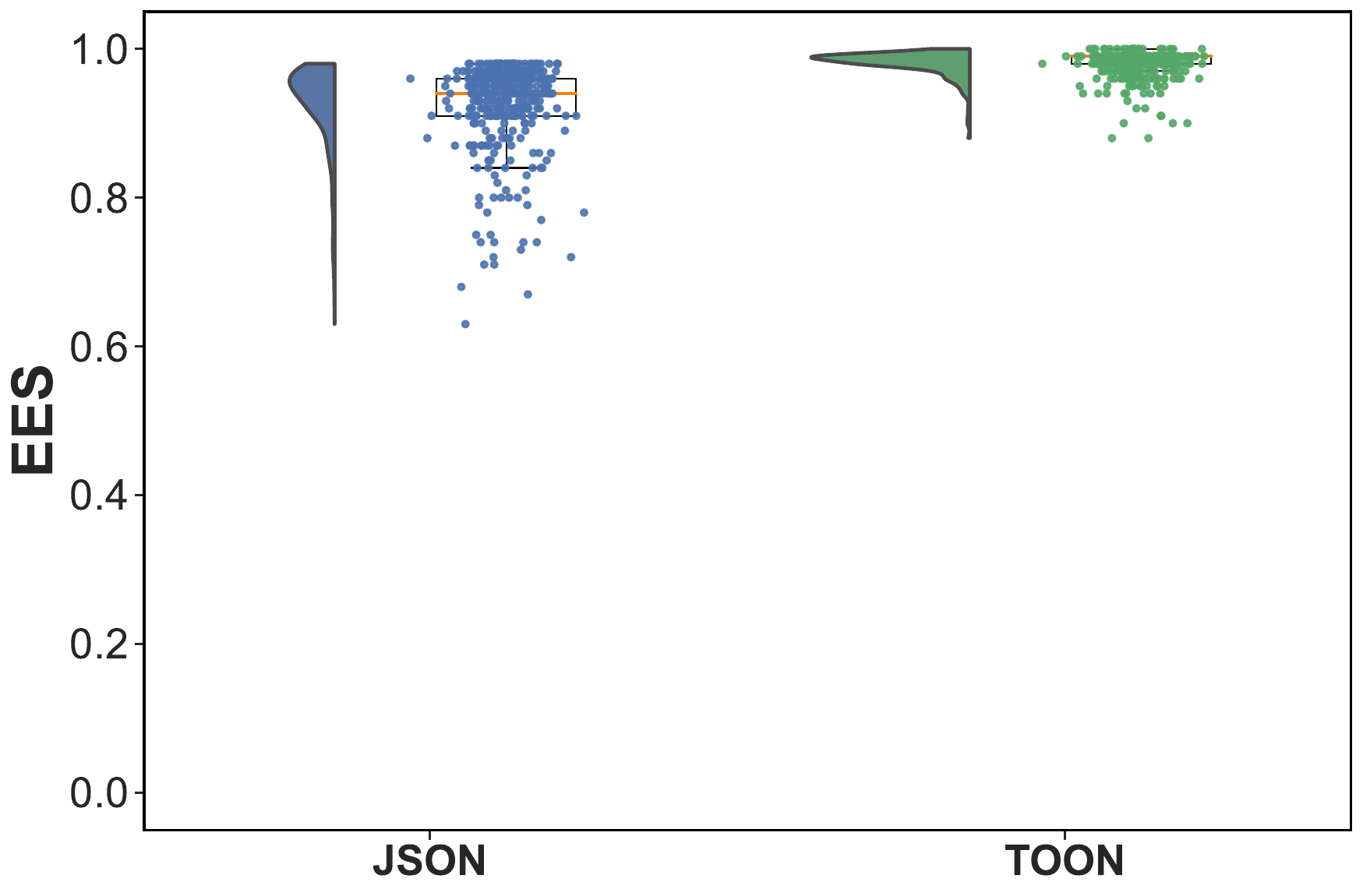}
        \caption{Environmental efficiency comparison (EES): JSON vs TOON.}
        \label{fig:img21_EES}
    \end{subfigure}\hfill
    \begin{subfigure}[t]{0.32\textwidth}
        \centering
        \includegraphics[width=\linewidth]{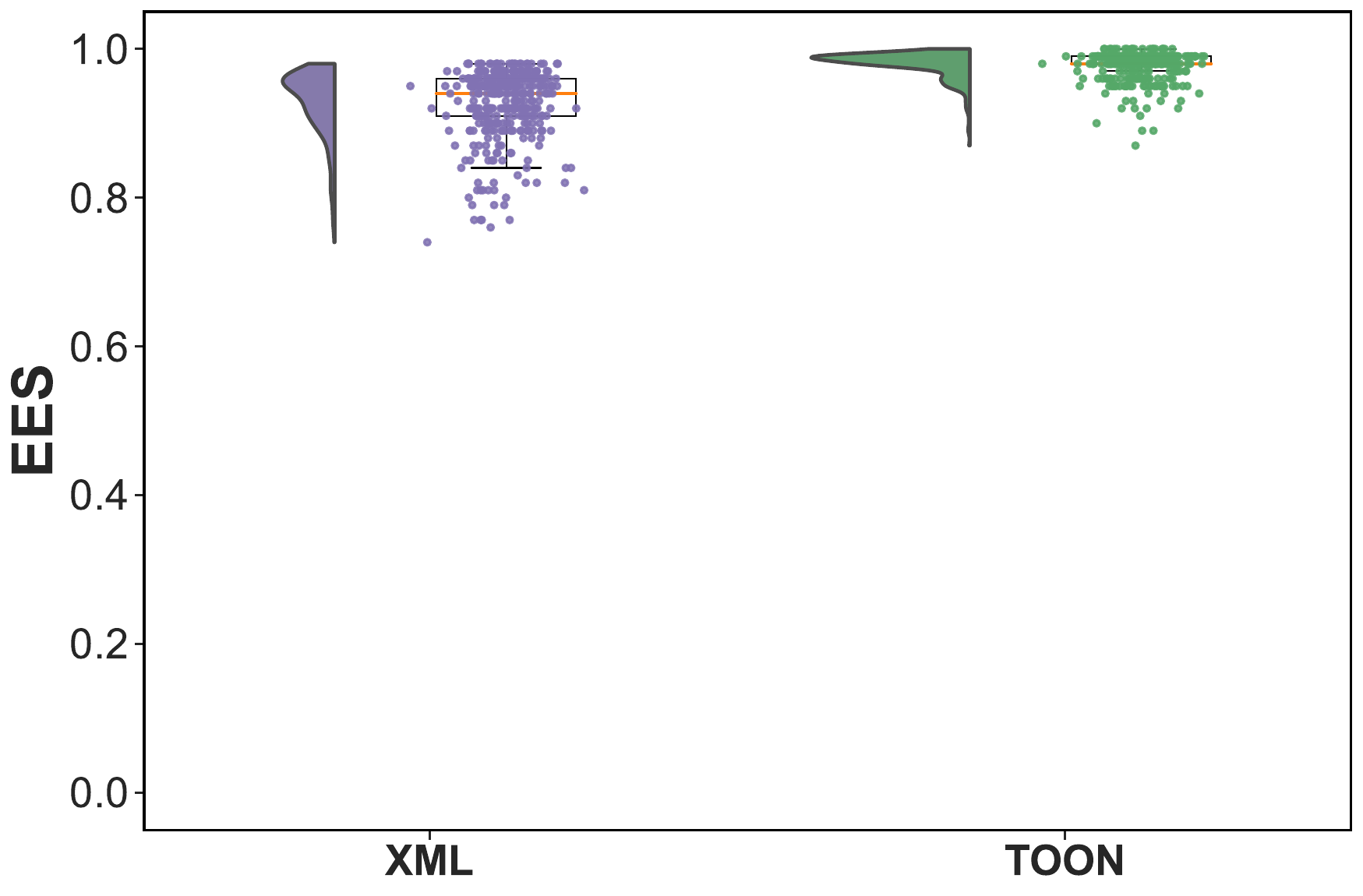}
        \caption{Environmental efficiency comparison (EES): XML vs TOON (no stat sign)}
        \label{fig:img22_EES}
    \end{subfigure}\hfill
    \begin{subfigure}[t]{0.32\textwidth}
        \centering
        \includegraphics[width=\linewidth]{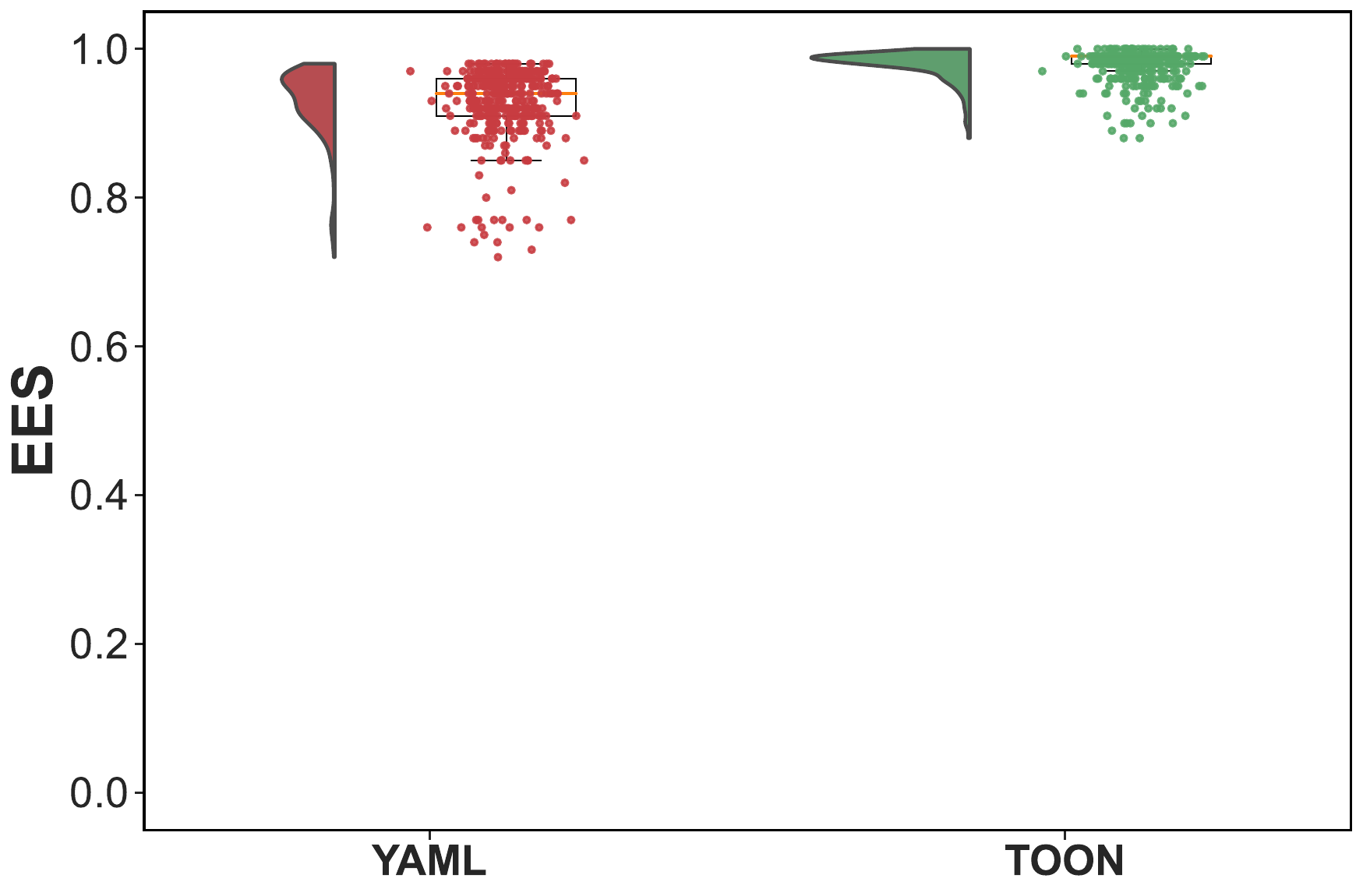}
        \caption{Environmental efficiency comparison (EES): YAML vs TOON (no stat sign)}
        \label{fig:img23_EES}
    \end{subfigure}
    \vspace{0.8em}
    \begin{subfigure}[t]{0.32\textwidth}
        \centering
        \includegraphics[width=\linewidth]{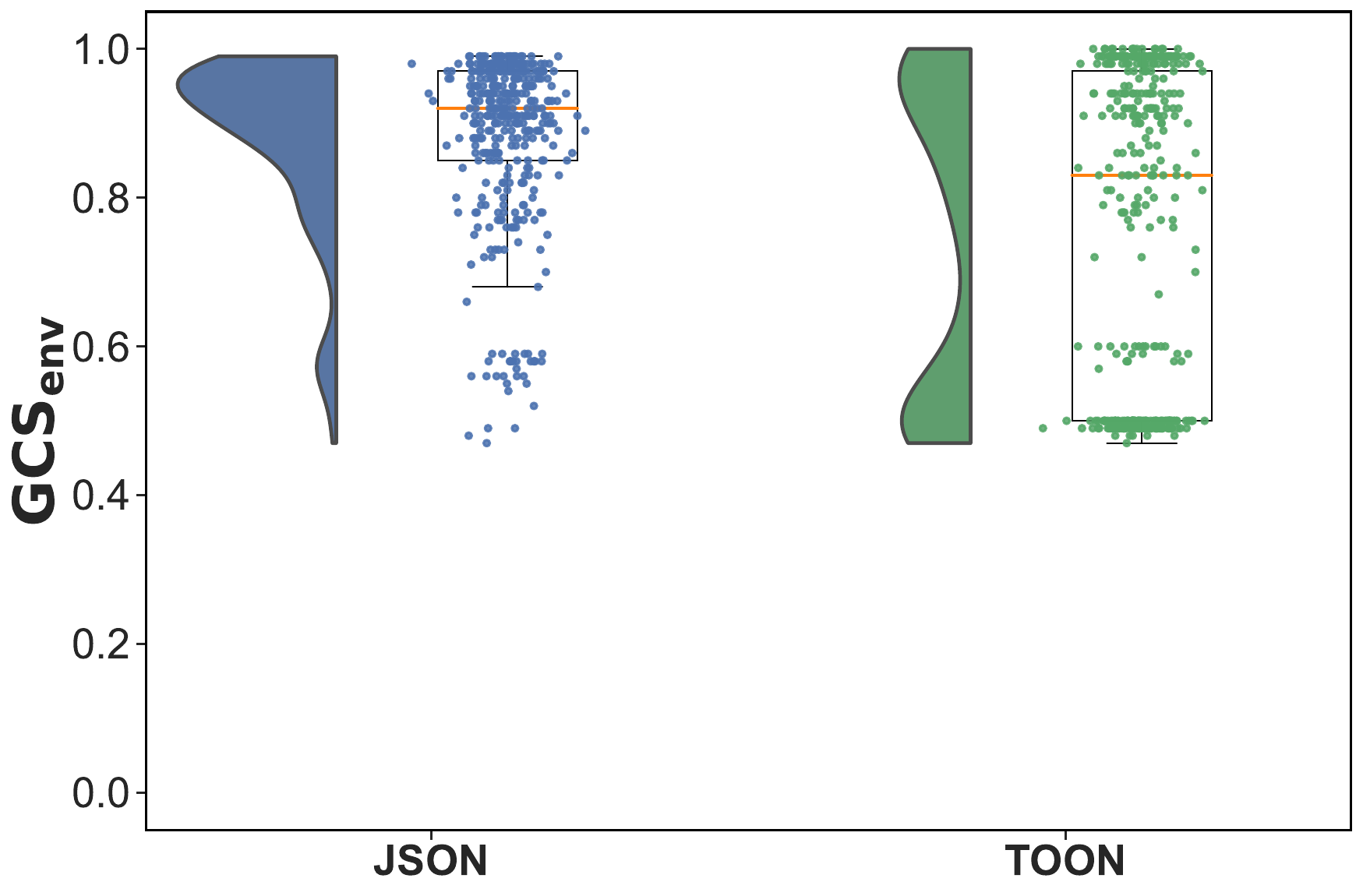}
        \caption{Environment-aware generation correctness ($GCS_{env}$): JSON vs TOON.}
        \label{fig:img31_GCS_ENV}
    \end{subfigure}\hfill
    \begin{subfigure}[t]{0.32\textwidth}
        \centering
        \includegraphics[width=\linewidth]{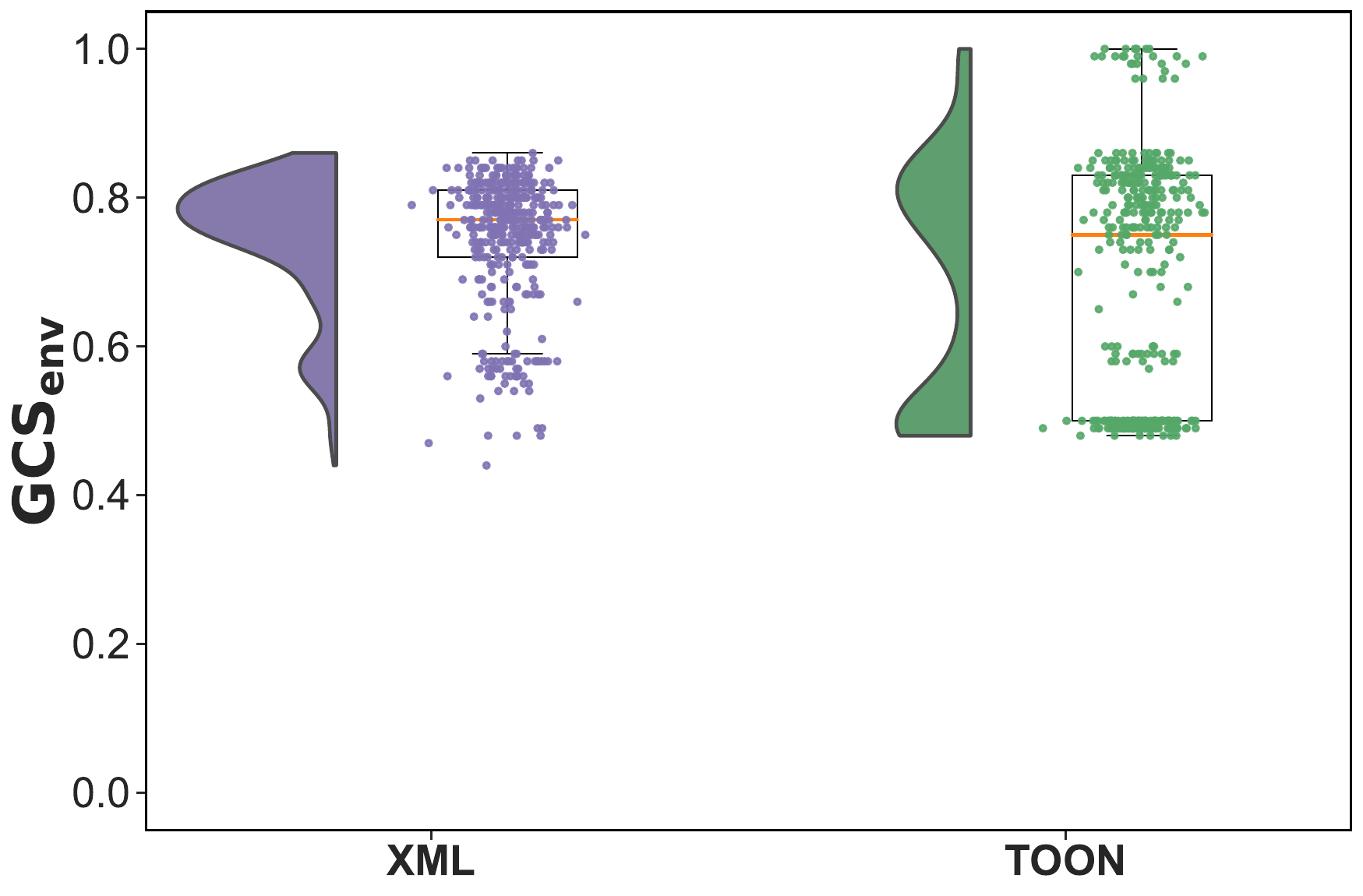}
        \caption{Environment-aware generation correctness ($GCS_{env}$): XML vs TOON.}
        \label{fig:img32_GCS_ENV}
    \end{subfigure}\hfill
    \begin{subfigure}[t]{0.32\textwidth}
        \centering
        \includegraphics[width=\linewidth]{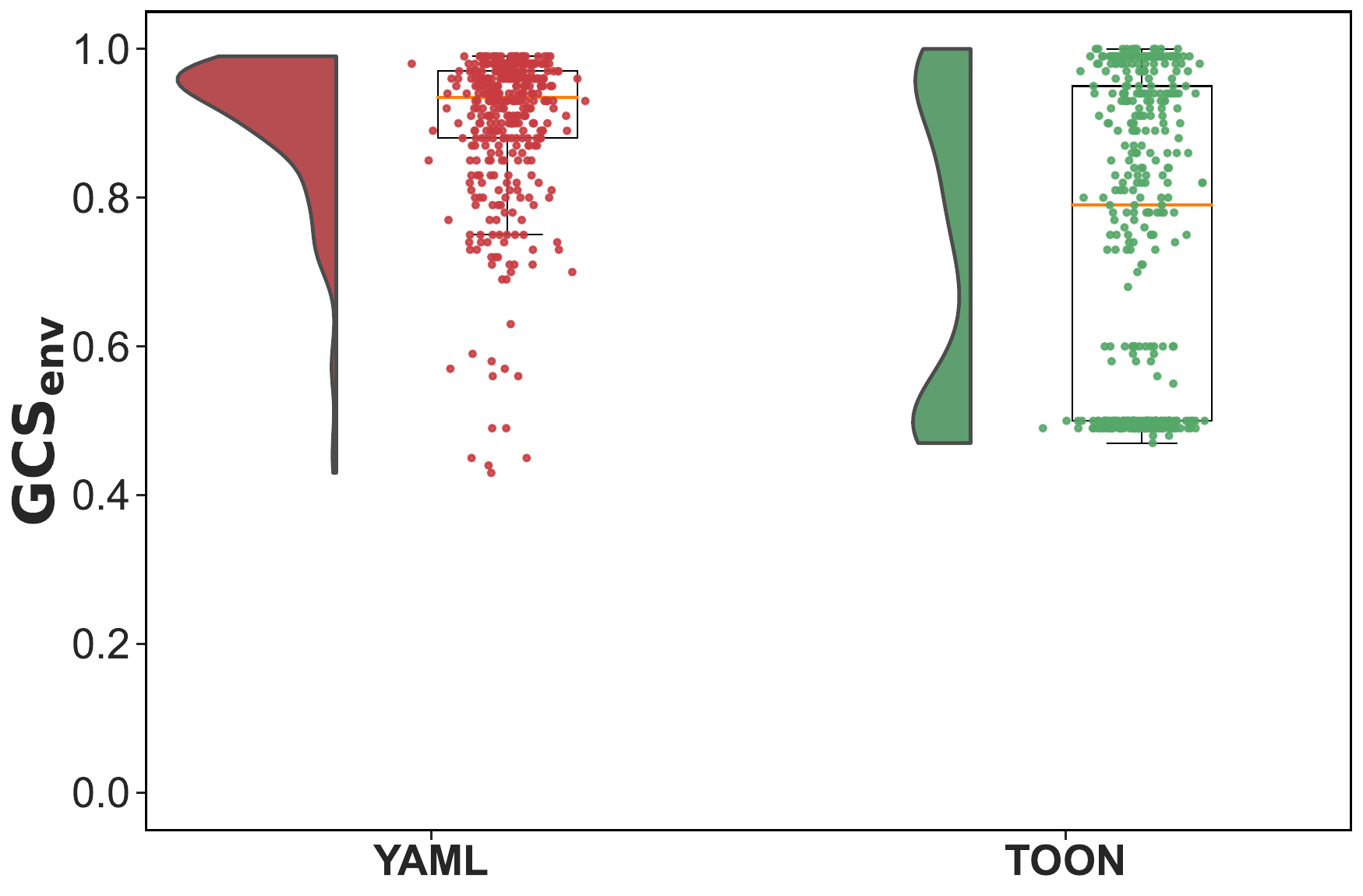}
        \caption{Environment-aware generation correctness ($GCS_{env}$): YAML vs TOON.}
        \label{fig:img33_GCS_ENV}
    \end{subfigure}
    \caption{Paired comparison between TOON and standard structured formats across correctness, efficiency, and environment-aware metrics.}
    \label{fig:grid3x3}
\end{figure*}

Overall, the aggregated results reveal a highly consistent trade-off between structural correctness and efficiency that persists across models. From an efficiency perspective, TOON systematically produces more compact outputs, as reflected by a substantially lower number of generated tokens ($N_T$) with respect to JSON, XML, and YAML (26.4\%, −49.4\%, and −15.3\%, respectively). This reduction in token count directly translates into markedly lower estimated carbon emissions (\textit{CE}) during the output generation phase (−82.9\%, −87.6\%, and −76.9\%), yielding consistently higher values of the Environmental Efficiency Score (\textit{EES}) across all comparisons (+5.8\%, +5.5\%, and +5.4\%), with TOON achieving near-ceiling \textit{EES} values ($\approx 0.98$) regardless of the baseline representation. These differences are statistically significant according to the Wilcoxon signed-rank test ($p < 0.05$) and are consistently reflected in the score distributions shown in Figures \ref{fig:img11_GCS}, \ref{fig:img12_GCS}, and \ref{fig:img13_GCS}.

In contrast, baseline formats consistently outperform TOON in terms of structural correctness. JSON, XML, and YAML achieve higher Render and Syntax Scores, resulting in substantially higher Generation Correctness Scores (\textit{GCS}) in all aggregate comparisons. This gap reflects TOON’s reduced robustness in strictly adhering to formal structural constraints (−38.9\%, −30.9\%, and −42.2\% with respect to JSON, XML, and YAML, respectively), which can be attributed to the fact that TOON is not natively supported by the evaluated LLMs and must be enforced exclusively through prompt-level instructions. The observed differences in \textit{GCS} are statistically significant under the Wilcoxon signed-rank test ($p < 0.05$, see Figures \ref{fig:img21_EES}, \ref{fig:img22_EES}, and \ref{fig:img23_EES}.

When correctness and environmental efficiency are jointly considered through the proposed environment-aware score $GCS_{env}$, baseline formats continue to achieve higher overall scores in the aggregate. Although TOON benefits from a strong advantage in environmental efficiency, these gains are insufficient to offset the loss in structural correctness under the adopted weighting scheme ($\gamma = 0.5$). Nevertheless, the gap between TOON and baseline formats is substantially reduced when moving from \textit{GCS} to $GCS_{env}$, highlighting the role of carbon-aware evaluation in moderating purely correctness-driven assessments \cite{pistolesisustainability}. Importantly, the differences observed in $GCS_{env}$ remain statistically significant ($p < 0.05$, see Figures \ref{fig:img31_GCS_ENV}, \ref{fig:img32_GCS_ENV}, and \ref{fig:img33_GCS_ENV}.

Finally, generation latency show the only notable deviation from otherwise uniform trends across formats and metrics: TOON exhibits slightly longer generation times compared to JSON and YAML, while achieving lower duration than XML. This behavior can be plausibly explained by the well-documented verbosity of XML representations \cite{nurseitov2009comparison}, which require explicit opening and closing tags and therefore induce substantially larger token sequences during generation. In our setting, this verbosity is reflected in the significantly higher $N_T$ observed for XML outputs (approximately 472 generated tokens on average), compared to consistently more compact outputs for the other formats (all below 300 tokens on average).

These results provide clear evidence of stable, model-agnostic trends: TOON consistently improves efficiency and environmental sustainability at the cost of structural correctness. These findings motivate a deeper, model-wise analysis in the following sections, where we examine how such trade-offs evolve across architectures and parameter scales.

\subsection{Model-wise Comparison of Structured Formats} \label{sec:results_modelwise}

To address \textbf{RQ2}, we conduct a model-wise analysis to examine how the trade-offs between structural correctness and environmental efficiency vary across model parameter scales. In the following subsections, we separately analyze \textit{(i)} structural correctness, to assess the extent to which increasing model capacity mitigates the lack of native support for novel formats such as TOON, \textit{(ii)} environmental impact, to evaluate how efficiency gains evolve across models of different sizes, and \textit{(iii)} an environment-aware evaluation that jointly accounts for correctness and sustainability, highlighting how these dimensions interact under a unified scoring framework.

\subsubsection{\textbf{Structural Correctness}} \label{sec:results_modelwise_correctness}

Table \ref{tab:gcs_per_model} reports \textit{GCS} scores across all evaluated LLMs for all considered output formats.

\begin{table*}[t]
    \centering
    \resizebox{0.7\textwidth}{!}{%
    \begin{tabular}{l|cc|cc|cc}
    \hline
    \textbf{Model} 
    & \textbf{JSON} & \textbf{TOON} 
    & \textbf{XML}  & \textbf{TOON} 
    & \textbf{YAML} & \textbf{TOON} \\
    \hline

    GPT-oss 20B
    & \makecell{\small \textbf{0.862}\\[-3pt]{\scriptsize(± 0.211)}}
    & \makecell{\small 0.793\\[-3pt]{\scriptsize(± 0.302)}}
    & \makecell{\small \textbf{0.609}\\[-3pt]{\scriptsize(± 0.119)}}
    & \makecell{\small 0.559\\[-3pt]{\scriptsize(± 0.224)}}
    & \makecell{\small \textbf{0.893}\\[-3pt]{\scriptsize(± 0.170)}}
    & \makecell{\small 0.827\\[-3pt]{\scriptsize(± 0.273)}} \\[5pt]

    GPT-oss 120B
    & \makecell{\small \textbf{0.910}\\[-3pt]{\scriptsize(± 0.128)}}
    & \makecell{\small 0.834\\[-3pt]{\scriptsize(± 0.257)}}
    & \makecell{\small 0.608\\[-3pt]{\scriptsize(± 0.122)}}
    & \makecell{\small \textbf{0.614}\\[-3pt]{\scriptsize(± 0.224)}}
    & \makecell{\small \textbf{0.907}\\[-3pt]{\scriptsize(± 0.137)}}
    & \makecell{\small 0.831\\[-3pt]{\scriptsize(± 0.241)}} \\
    \hline

    Gemma 3 4B
    & \makecell{\small \textbf{0.779}\\[-3pt]{\scriptsize(± 0.259)}}
    & \makecell{\small 0.045\\[-3pt]{\scriptsize(± 0.156)}}
    & \makecell{\small \textbf{0.579}\\[-3pt]{\scriptsize(± 0.155)}}
    & \makecell{\small 0.157\\[-3pt]{\scriptsize(± 0.273)}}
    & \makecell{\small \textbf{0.818}\\[-3pt]{\scriptsize(± 0.188)}}
    & \makecell{\small 0.051\\[-3pt]{\scriptsize(± 0.163)}} \\[5pt]

    Gemma 3 12B
    & \makecell{\small \textbf{0.875}\\[-3pt]{\scriptsize(± 0.189)}}
    & \makecell{\small 0.278\\[-3pt]{\scriptsize(± 0.397)}}
    & \makecell{\small \textbf{0.580}\\[-3pt]{\scriptsize(± 0.145)}}
    & \makecell{\small 0.117\\[-3pt]{\scriptsize(± 0.261)}}
    & \makecell{\small \textbf{0.906}\\[-3pt]{\scriptsize(± 0.140)}}
    & \makecell{\small 0.233\\[-3pt]{\scriptsize(± 0.387)}} \\[5pt]

    Gemma 3 27B
    & \makecell{\small \textbf{0.871}\\[-3pt]{\scriptsize(± 0.189)}}
    & \makecell{\small 0.569\\[-3pt]{\scriptsize(± 0.438)}}
    & \makecell{\small \textbf{0.488}\\[-3pt]{\scriptsize(± 0.204)}}
    & \makecell{\small 0.417\\[-3pt]{\scriptsize(± 0.346)}}
    & \makecell{\small \textbf{0.868}\\[-3pt]{\scriptsize(± 0.215)}}
    & \makecell{\small 0.484\\[-3pt]{\scriptsize(± 0.435)}} \\
    \hline

    Mistral 7B
    & \makecell{\small \textbf{0.848}\\[-3pt]{\scriptsize(± 0.224)}}
    & \makecell{\small 0.435\\[-3pt]{\scriptsize(± 0.417)}}
    & \makecell{\small \textbf{0.513}\\[-3pt]{\scriptsize(± 0.203)}}
    & \makecell{\small 0.381\\[-3pt]{\scriptsize(± 0.351)}}
    & \makecell{\small \textbf{0.903}\\[-3pt]{\scriptsize(± 0.142)}}
    & \makecell{\small 0.355\\[-3pt]{\scriptsize(± 0.399)}} \\
    \hline

    Llama 3.3 70B
    & \makecell{\small 0.808\\[-3pt]{\scriptsize(± 0.262)}}
    & \makecell{\small \textbf{0.819}\\[-3pt]{\scriptsize(± 0.294)}}
    & \makecell{\small 0.596\\[-3pt]{\scriptsize(± 0.146)}}
    & \makecell{\small \textbf{0.611}\\[-3pt]{\scriptsize(± 0.243)}}
    & \makecell{\small \textbf{0.828}\\[-3pt]{\scriptsize(± 0.281)}}
    & \makecell{\small 0.786\\[-3pt]{\scriptsize(± 0.305)}} \\
    \hline

    Qwen 3 4B
    & \makecell{\small \textbf{0.769}\\[-3pt]{\scriptsize(± 0.313)}}
    & \makecell{\small 0.334\\[-3pt]{\scriptsize(± 0.413)}}
    & \makecell{\small \textbf{0.529}\\[-3pt]{\scriptsize(± 0.202)}}
    & \makecell{\small 0.253\\[-3pt]{\scriptsize(± 0.325)}}
    & \makecell{\small \textbf{0.856}\\[-3pt]{\scriptsize(± 0.239)}}
    & \makecell{\small 0.463\\[-3pt]{\scriptsize(± 0.369)}} \\

    \hline
    \end{tabular}}
    \caption{Per-model \textit{GCS} results for each format. Mean shown above and standard deviation in parentheses below.}
    \label{tab:gcs_per_model}
\end{table*}

\begin{figure*}[t]
    \centering
    \begin{subfigure}[t]{0.32\textwidth}
        \centering
        \includegraphics[width=\linewidth]{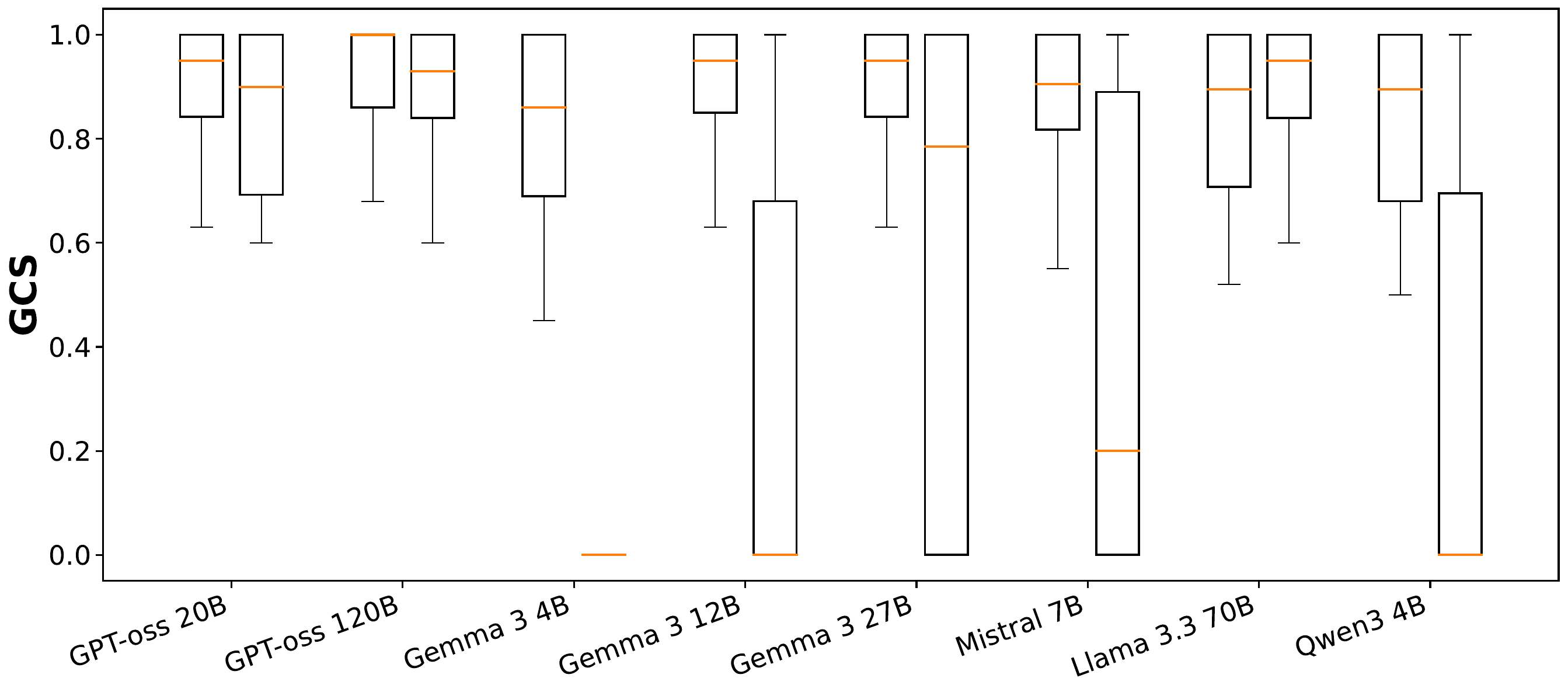}
        \label{fig:img11_BOX_GCS}
    \end{subfigure}\hfill
    \begin{subfigure}[t]{0.32\textwidth}
        \centering
        \includegraphics[width=\linewidth]{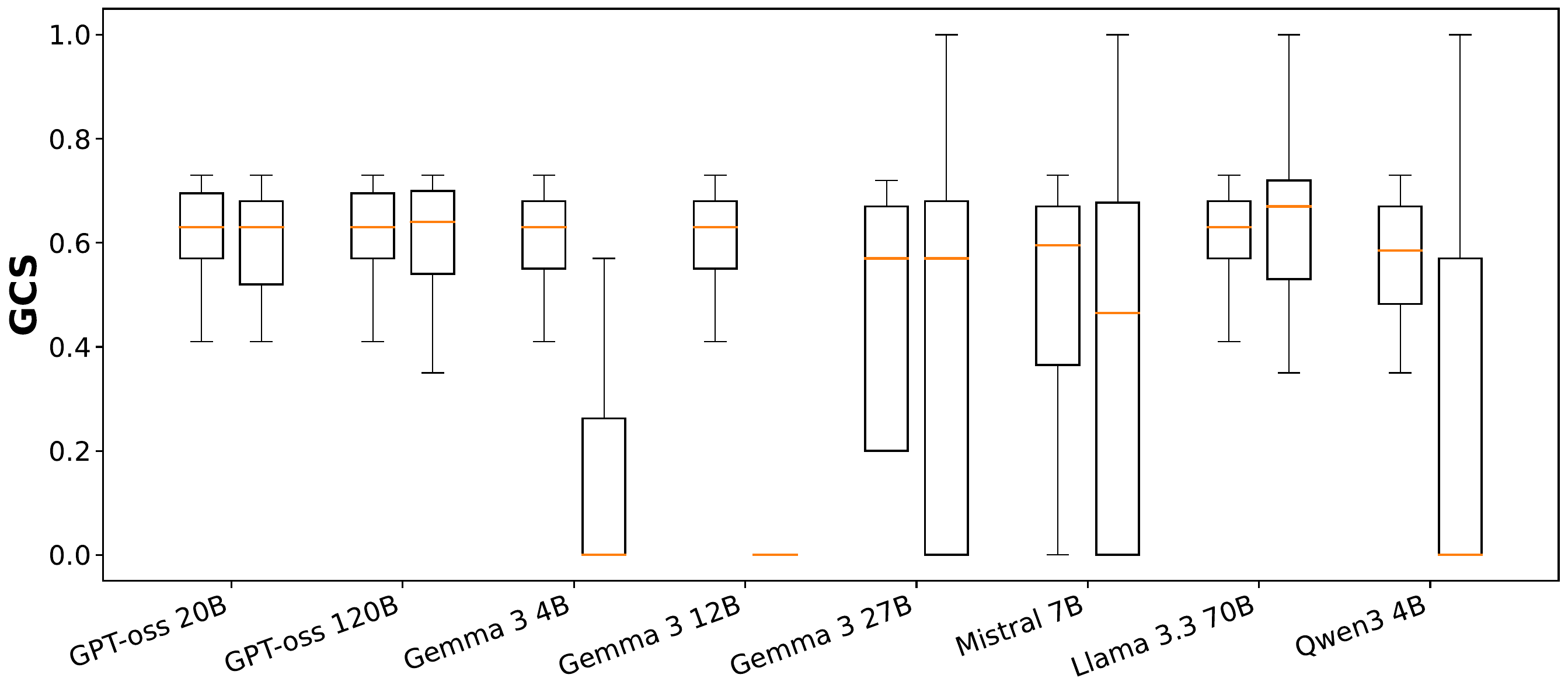}
        \label{fig:img12_BOX_GCS}
    \end{subfigure}\hfill
    \begin{subfigure}[t]{0.32\textwidth}
        \centering
        \includegraphics[width=\linewidth]{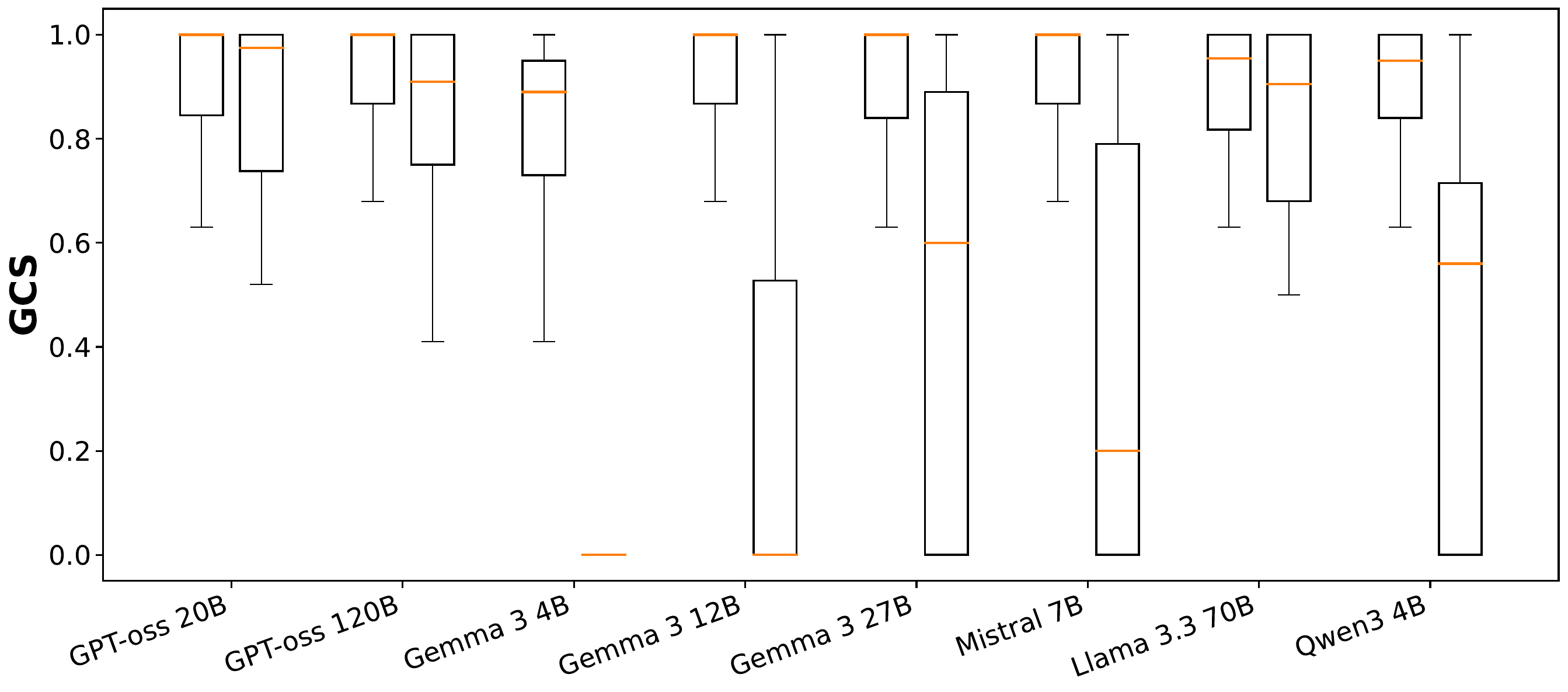}
        \label{fig:img13_BOX_GCS}
    \end{subfigure}
    \vspace{0.8em}
    \begin{subfigure}[t]{0.32\textwidth}
        \centering
        \includegraphics[width=\linewidth]{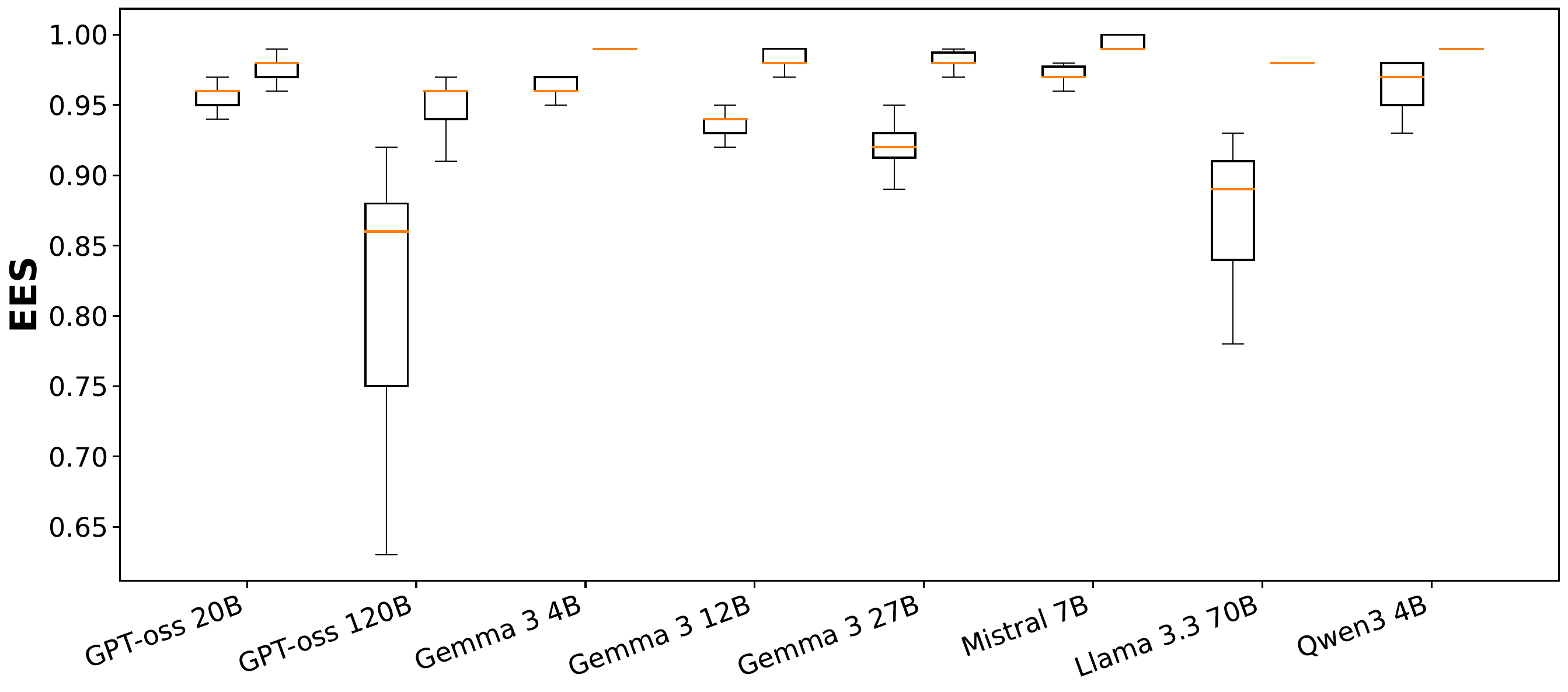}
        \label{fig:img21_BOX_EES}
    \end{subfigure}\hfill
    \begin{subfigure}[t]{0.32\textwidth}
        \centering
        \includegraphics[width=\linewidth]{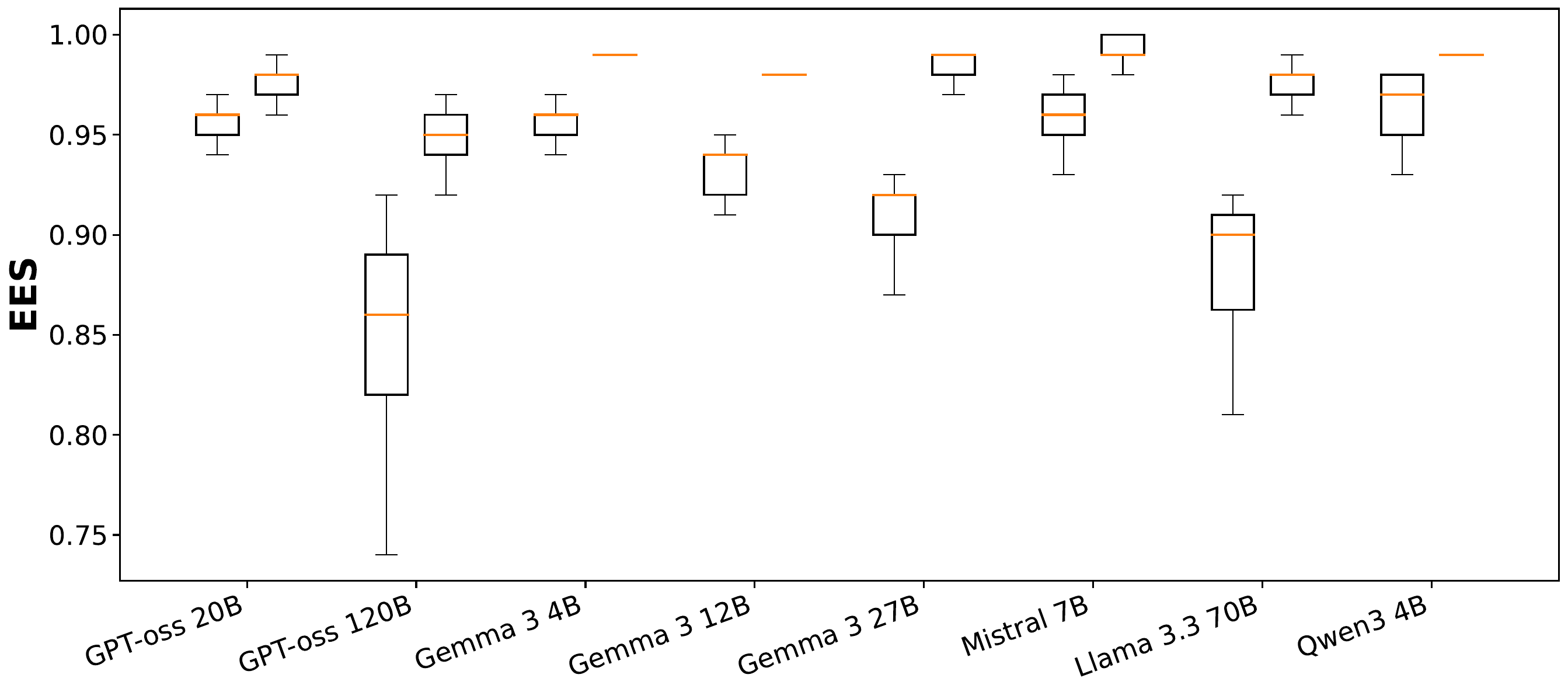}
        \label{fig:img22_BOX_EES}
    \end{subfigure}\hfill
    \begin{subfigure}[t]{0.32\textwidth}
        \centering
        \includegraphics[width=\linewidth]{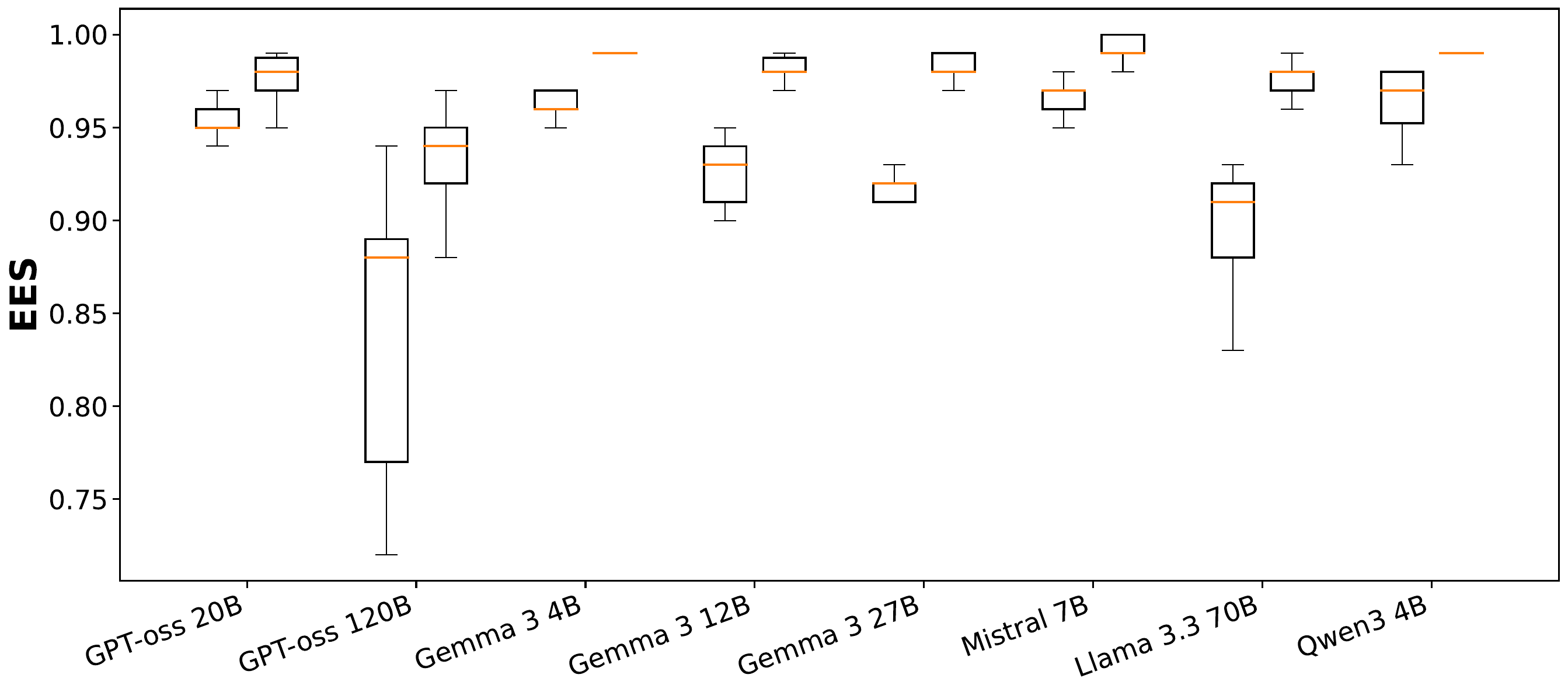}
        \label{fig:img23_BOX_EES}
    \end{subfigure}
    \vspace{0.8em}
    \begin{subfigure}[t]{0.32\textwidth}
        \centering
        \includegraphics[width=\linewidth]{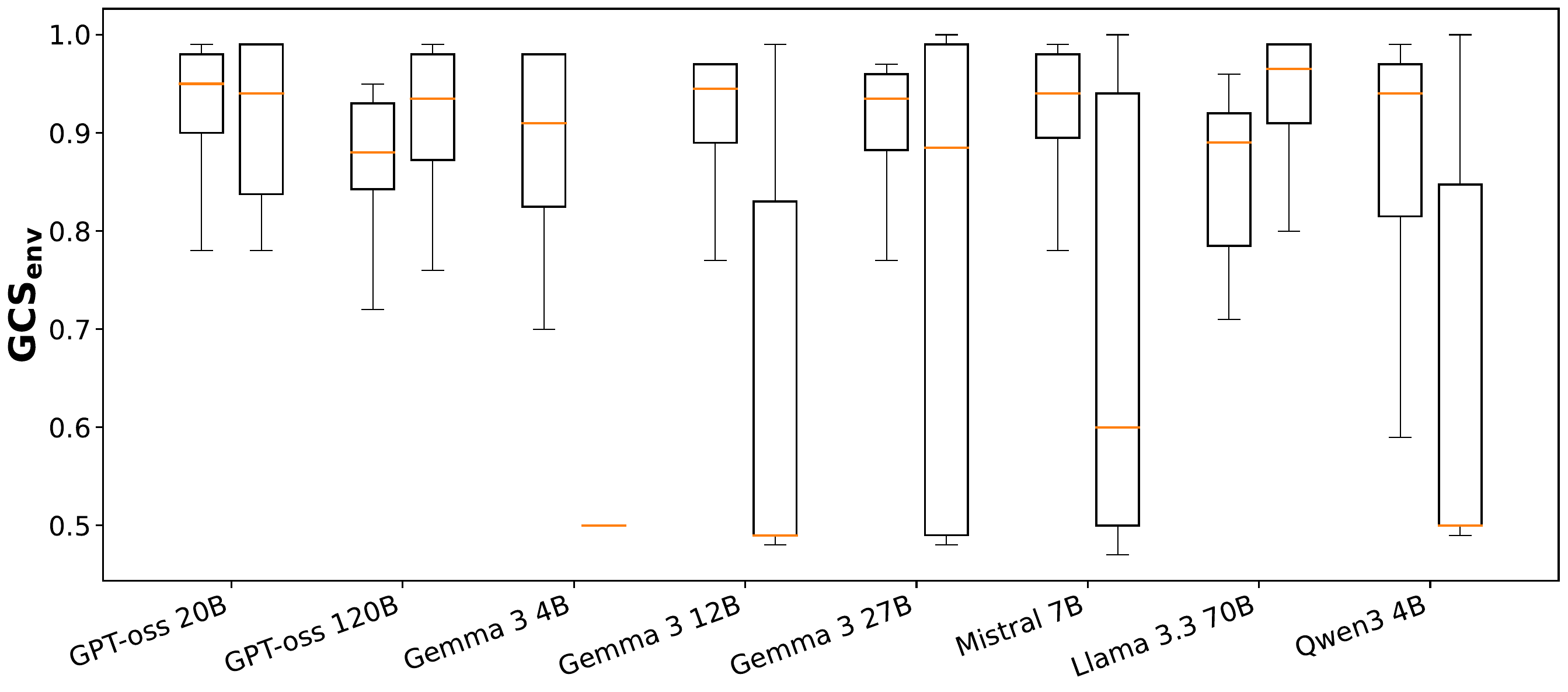}
        \caption{JSON vs TOON}
        \label{fig:img31_BOX_GCS_ENV}
    \end{subfigure}\hfill
    \begin{subfigure}[t]{0.32\textwidth}
        \centering
        \includegraphics[width=\linewidth]{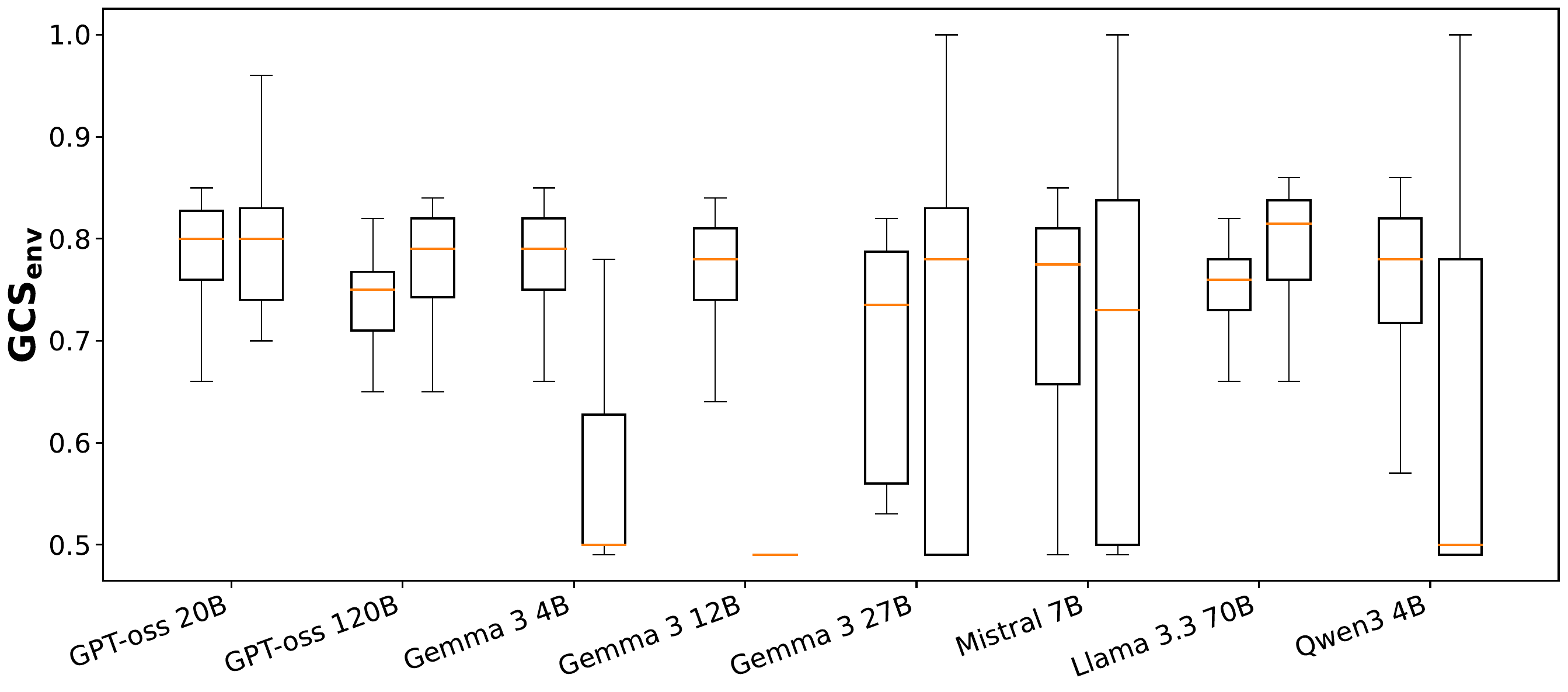}
        \caption{XML vs TOON}
        \label{fig:img32_BOX_GCS_ENV}
    \end{subfigure}\hfill
    \begin{subfigure}[t]{0.32\textwidth}
        \centering
        \includegraphics[width=\linewidth]{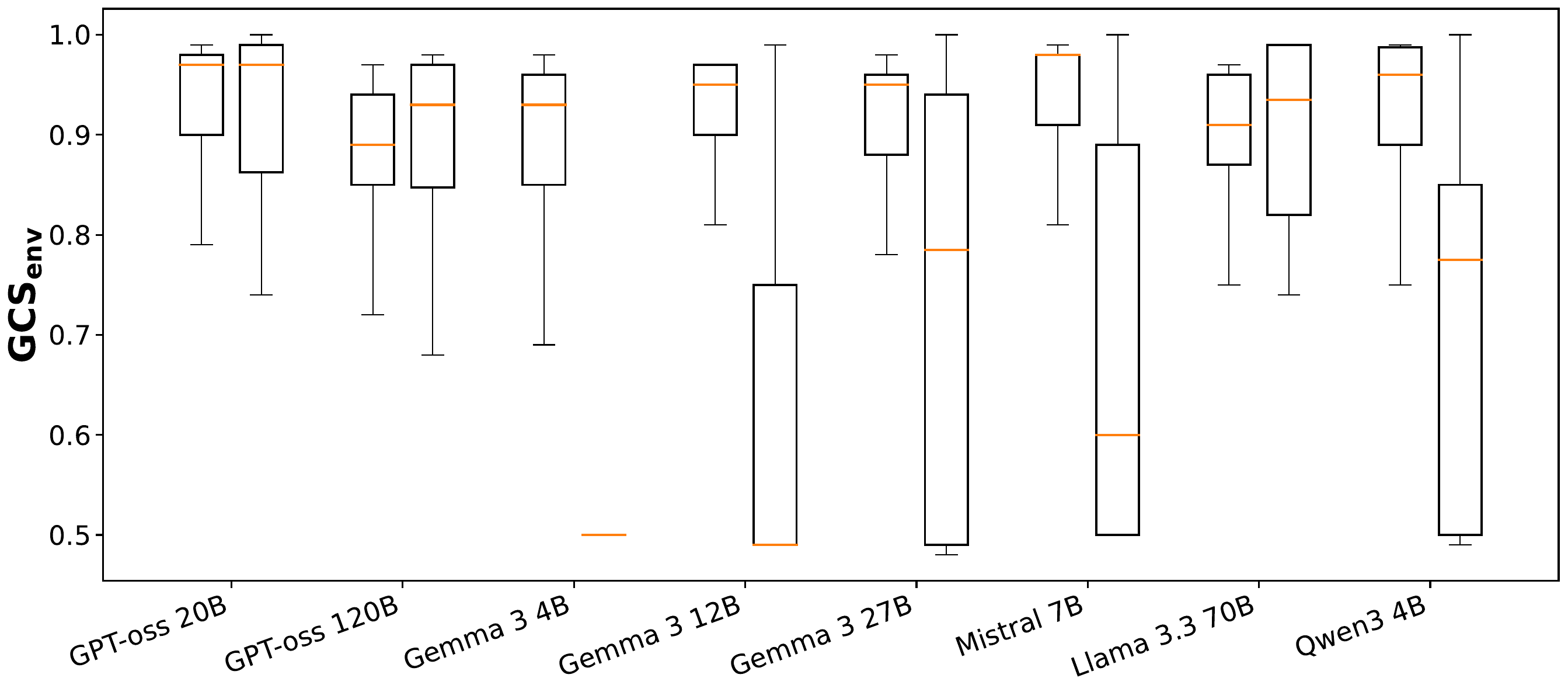}
        \caption{YAML vs TOON}
        \label{fig:img33_BOX_GCS_ENV}
    \end{subfigure}
    \caption{Titolo generale della griglia 3$\times$3.}
    \label{fig:grid3x3}
\end{figure*}

An emerging and consistent trend is that larger models consistently reduce the structural correctness gap between baseline formats and TOON. While smaller and mid-sized models exhibit substantial degradation in \textit{GCS} when generating TOON outputs, this gap progressively narrows as model scale increases and, in some cases, is fully eliminated. In particular, the Wilcoxon signed-rank test indicates no statistically significant differences ($p > 0.05$) between XML and TOON in Gemma 3 27B, GPT-oss 20B and GPT-oss 120B. More strikingly, for Llama 3.3 70B, no statistically significant differences are observed when comparing TOON against any baseline format. All remaining model–format comparisons yield statistically significant differences ($p < 0.05$), with baseline formats consistently achieving higher structural correctness than TOON. In extreme cases, such as Gemma 3 4B, TOON generation exhibits near-complete failure in satisfying structural constraints.

Collectively, these results indicate that structural correctness differences are both systematic and scale-dependent, rather than attributable to noise or isolated model-specific behaviors. In particular, the observed trends suggest that model capacity plays a central role in mitigating the absence of native format support, with larger models exhibiting a markedly higher ability to internalize and apply previously unseen structural constraints. As a result, the practical feasibility of compact structured representations such as TOON appears tightly coupled with advances in LLM capacity and training, rather than being solely determined by the design of the format itself.

\subsubsection{\textbf{Environmental Impact}} \label{sec:results_modelwise_efficiency}

Table \ref{tab:eff_env_per_model} reports per-model \textit{EES} for TOON and all baseline formats. This analysis complements the structural correctness evaluation by examining how environmental efficiency varies across model families and parameter scales.

\begin{table*}[t]
    \centering
    \resizebox{0.7\textwidth}{!}{%
    \begin{tabular}{l|cc|cc|cc}
    \hline
    \textbf{Model} 
    & \textbf{JSON} & \textbf{TOON} 
    & \textbf{XML}  & \textbf{TOON} 
    & \textbf{YAML} & \textbf{TOON} \\
    \hline

    GPT-oss 20B
    & \makecell{\small 0.956\\[-3pt]{\scriptsize(± 0.008)}}
    & \makecell{\small \textbf{0.975}\\[-3pt]{\scriptsize(± 0.010)}}
    & \makecell{\small 0.959\\[-3pt]{\scriptsize(± 0.009)}}
    & \makecell{\small \textbf{0.977}\\[-3pt]{\scriptsize(± 0.011)}}
    & \makecell{\small 0.952\\[-3pt]{\scriptsize(± 0.011)}}
    & \makecell{\small \textbf{0.976}\\[-3pt]{\scriptsize(± 0.011)}} \\[5pt]

    GPT-oss 120B
    & \makecell{\small 0.825\\[-3pt]{\scriptsize(± 0.077)}}
    & \makecell{\small \textbf{0.947}\\[-3pt]{\scriptsize(± 0.024)}}
    & \makecell{\small 0.854\\[-3pt]{\scriptsize(± 0.046)}}
    & \makecell{\small \textbf{0.944}\\[-3pt]{\scriptsize(± 0.021)}}
    & \makecell{\small 0.846\\[-3pt]{\scriptsize(± 0.064)}}
    & \makecell{\small \textbf{0.936}\\[-3pt]{\scriptsize(± 0.024)}} \\
    \hline

    Gemma 3 4B
    & \makecell{\small 0.962\\[-3pt]{\scriptsize(± 0.005)}}
    & \makecell{\small \textbf{0.990}\\[-3pt]{\scriptsize(± 0.001)}}
    & \makecell{\small 0.958\\[-3pt]{\scriptsize(± 0.007)}}
    & \makecell{\small \textbf{0.990}\\[-3pt]{\scriptsize(± 0.000)}}
    & \makecell{\small 0.963\\[-3pt]{\scriptsize(± 0.005)}}
    & \makecell{\small \textbf{0.990}\\[-3pt]{\scriptsize(± 0.000)}} \\[5pt]

    Gemma 3 12B
    & \makecell{\small 0.932\\[-3pt]{\scriptsize(± 0.012)}}
    & \makecell{\small \textbf{0.984}\\[-3pt]{\scriptsize(± 0.007)}}
    & \makecell{\small 0.933\\[-3pt]{\scriptsize(± 0.012)}}
    & \makecell{\small \textbf{0.978}\\[-3pt]{\scriptsize(± 0.006)}}
    & \makecell{\small 0.925\\[-3pt]{\scriptsize(± 0.015)}}
    & \makecell{\small \textbf{0.982}\\[-3pt]{\scriptsize(± 0.006)}} \\[5pt]

    Gemma 3 27B
    & \makecell{\small 0.924\\[-3pt]{\scriptsize(± 0.013)}}
    & \makecell{\small \textbf{0.982}\\[-3pt]{\scriptsize(± 0.005)}}
    & \makecell{\small 0.912\\[-3pt]{\scriptsize(± 0.016)}}
    & \makecell{\small \textbf{0.987}\\[-3pt]{\scriptsize(± 0.005)}}
    & \makecell{\small 0.918\\[-3pt]{\scriptsize(± 0.012)}}
    & \makecell{\small \textbf{0.984}\\[-3pt]{\scriptsize(± 0.007)}} \\
    \hline

    Mistral 7B
    & \makecell{\small 0.970\\[-3pt]{\scriptsize(± 0.008)}}
    & \makecell{\small \textbf{0.993}\\[-3pt]{\scriptsize(± 0.009)}}
    & \makecell{\small 0.957\\[-3pt]{\scriptsize(± 0.023)}}
    & \makecell{\small \textbf{0.993}\\[-3pt]{\scriptsize(± 0.007)}}
    & \makecell{\small 0.961\\[-3pt]{\scriptsize(± 0.030)}}
    & \makecell{\small \textbf{0.993}\\[-3pt]{\scriptsize(± 0.012)}} \\
    \hline

    Llama 3.3 70B
    & \makecell{\small 0.875\\[-3pt]{\scriptsize(± 0.045)}}
    & \makecell{\small \textbf{0.980}\\[-3pt]{\scriptsize(± 0.009)}}
    & \makecell{\small 0.884\\[-3pt]{\scriptsize(± 0.040)}}
    & \makecell{\small \textbf{0.975}\\[-3pt]{\scriptsize(± 0.016)}}
    & \makecell{\small 0.899\\[-3pt]{\scriptsize(± 0.029)}}
    & \makecell{\small \textbf{0.974}\\[-3pt]{\scriptsize(± 0.013)}} \\
    \hline

    Qwen 3 4B
    & \makecell{\small 0.965\\[-3pt]{\scriptsize(± 0.016)}}
    & \makecell{\small \textbf{0.991}\\[-3pt]{\scriptsize(± 0.003)}}
    & \makecell{\small 0.965\\[-3pt]{\scriptsize(± 0.019)}}
    & \makecell{\small \textbf{0.989}\\[-3pt]{\scriptsize(± 0.003)}}
    & \makecell{\small 0.963\\[-3pt]{\scriptsize(± 0.022)}}
    & \makecell{\small \textbf{0.990}\\[-3pt]{\scriptsize(± 0.002)}} \\
    \hline

    \end{tabular}}
    \caption{Per-model \textit{EES} results for each format. Mean shown above and standard deviation in parentheses below.}
    \label{tab:eff_env_per_model}
\end{table*}

A clear emerging trend is that larger models are associated with higher absolute emissions, as reflected by lower \textit{EES} values when moving from smaller to larger models. This behavior is expected, as larger models require substantially greater computational resources during decoding.

Across all evaluated models and formats, TOON consistently achieves higher \textit{EES} values than the corresponding baseline representations, indicating superior environmental efficiency. This improvement is remarkably stable across model families and parameter scales, including settings where TOON performs poorly in terms of structural correctness (see Section \ref{sec:results_modelwise_correctness}). The gains in \textit{EES} can be directly attributed to TOON’s more compact output representations, which systematically reduce the number of generated tokens and, consequently, the total carbon emissions associated with output generation. Wilcoxon signed-rank tests yield statistically significant differences in \textit{EES} ($p < 0.05$) across every evaluated model and format.

\begin{figure}[t]
    \centering
    \includegraphics[width=\linewidth]{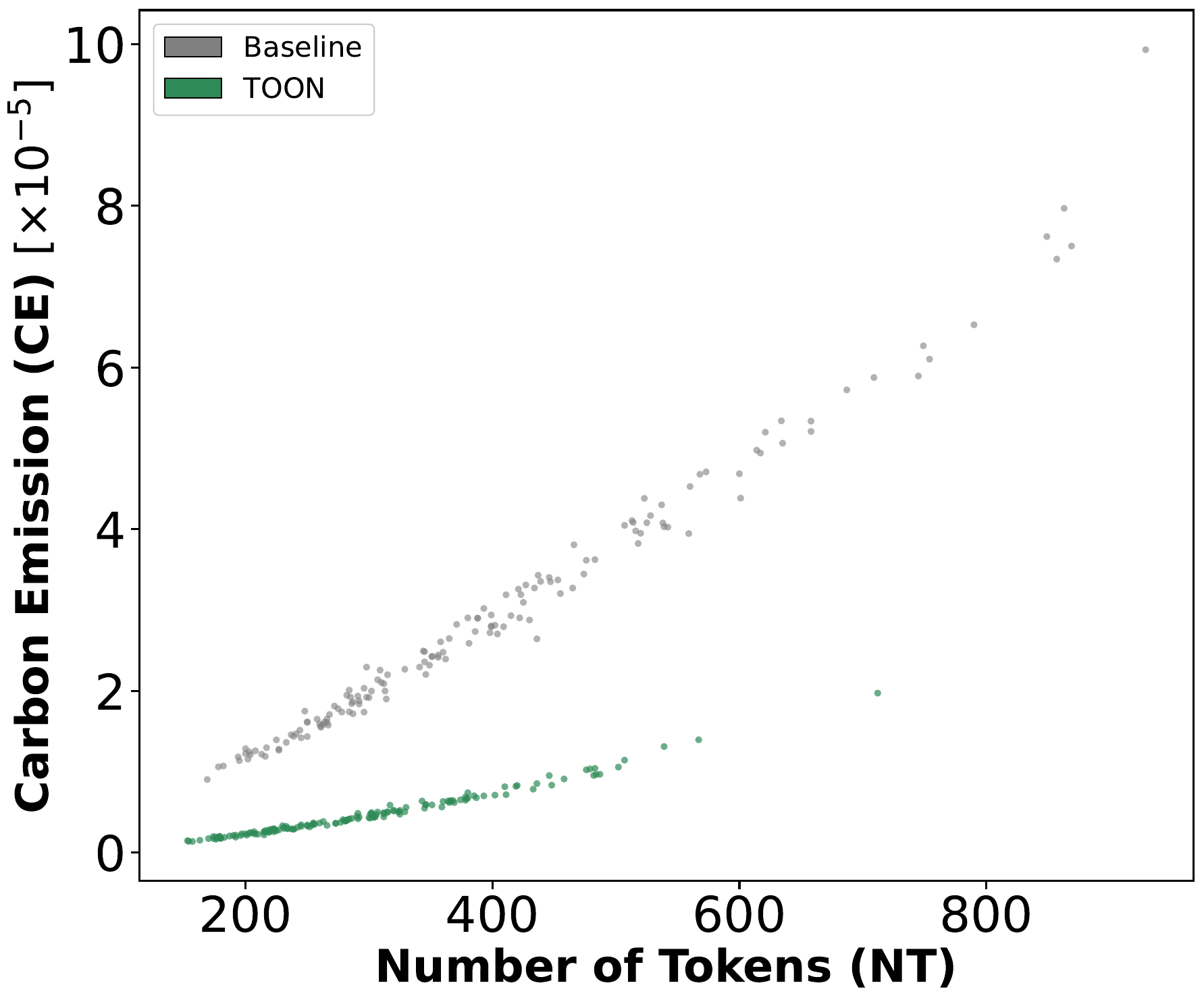}
    \caption{Token efficiency vs.\ carbon emissions across models and formats (e.g., $N_T$ vs.\ CE). Colors indicate formats; markers indicate models.}
    \label{fig:scatter_tokens_ce}
\end{figure}

The relationship between representational compactness and environmental impact is further illustrated in Figure \ref{fig:scatter_tokens_ce}, which reports a scatter plot of the number of generated tokens ($N_T$) versus estimated carbon emissions (\textit{CE}) for baseline formats and TOON. For clarity and readability, we report a single representative visualization corresponding to the Gemma 3 4B model, with the remaining models displaying the same qualitative trend. The figure highlights a clear correlation between token count and emissions, confirming that reductions in output length directly translate into lower environmental cost during decoding with respect to standard structured formats.

Interestingly, unlike structural correctness, environmental efficiency exhibits limited sensitivity to model scale: while absolute emissions increase for larger models, the relative advantage of TOON over JSON, XML, and YAML remains largely unchanged. This suggests that, from an environmental standpoint, the benefits of compact structured representations are largely orthogonal to advances in model capacity and persist even as models grow substantially larger.

\subsubsection{\textbf{Environment-Aware Evaluation}} \label{sec:results_envaware}

This analysis integrates the findings of the previous subsections and assesses how incorporating sustainability considerations affects the comparative evaluation of structured output formats.

\begin{table*}[t]
    \centering
    \resizebox{0.7\textwidth}{!}{%
    \begin{tabular}{l|cc|cc|cc}
    \hline
    \textbf{Model} 
    & \textbf{JSON} & \textbf{TOON} 
    & \textbf{XML}  & \textbf{TOON} 
    & \textbf{YAML} & \textbf{TOON} \\
    \hline

    GPT-oss 20B
    & \makecell{\small \textbf{0.909}\\[-3pt]{\scriptsize(± 0.107)}}
    & \makecell{\small 0.884\\[-3pt]{\scriptsize(± 0.153)}}
    & \makecell{\small \textbf{0.784}\\[-3pt]{\scriptsize(± 0.061)}}
    & \makecell{\small 0.768\\[-3pt]{\scriptsize(± 0.111)}}
    & \makecell{\small \textbf{0.924}\\[-3pt]{\scriptsize(± 0.087)}}
    & \makecell{\small 0.901\\[-3pt]{\scriptsize(± 0.139)}} \\[5pt]

    GPT-oss 120B
    & \makecell{\small 0.868\\[-3pt]{\scriptsize(± 0.074)}}
    & \makecell{\small \textbf{0.890}\\[-3pt]{\scriptsize(± 0.128)}}
    & \makecell{\small 0.730\\[-3pt]{\scriptsize(± 0.070)}}
    & \makecell{\small \textbf{0.780}\\[-3pt]{\scriptsize(± 0.110)}}
    & \makecell{\small 0.876\\[-3pt]{\scriptsize(± 0.079)}}
    & \makecell{\small \textbf{0.883}\\[-3pt]{\scriptsize(± 0.121)}} \\
    \hline

    Gemma 3 4B
    & \makecell{\small \textbf{0.869}\\[-3pt]{\scriptsize(± 0.130)}}
    & \makecell{\small 0.521\\[-3pt]{\scriptsize(± 0.077)}}
    & \makecell{\small \textbf{0.768}\\[-3pt]{\scriptsize(± 0.077)}}
    & \makecell{\small 0.575\\[-3pt]{\scriptsize(± 0.135)}}
    & \makecell{\small \textbf{0.891}\\[-3pt]{\scriptsize(± 0.094)}}
    & \makecell{\small 0.523\\[-3pt]{\scriptsize(± 0.082)}} \\[5pt]

    Gemma 3 12B
    & \makecell{\small \textbf{0.904}\\[-3pt]{\scriptsize(± 0.095)}}
    & \makecell{\small 0.629\\[-3pt]{\scriptsize(± 0.199)}}
    & \makecell{\small \textbf{0.756}\\[-3pt]{\scriptsize(± 0.073)}}
    & \makecell{\small 0.548\\[-3pt]{\scriptsize(± 0.131)}}
    & \makecell{\small \textbf{0.916}\\[-3pt]{\scriptsize(± 0.070)}}
    & \makecell{\small 0.606\\[-3pt]{\scriptsize(± 0.194)}} \\[5pt]

    Gemma 3 27B
    & \makecell{\small \textbf{0.897}\\[-3pt]{\scriptsize(± 0.096)}}
    & \makecell{\small 0.776\\[-3pt]{\scriptsize(± 0.220)}}
    & \makecell{\small \textbf{0.700}\\[-3pt]{\scriptsize(± 0.103)}}
    & \makecell{\small 0.700\\[-3pt]{\scriptsize(± 0.173)}}
    & \makecell{\small \textbf{0.894}\\[-3pt]{\scriptsize(± 0.107)}}
    & \makecell{\small 0.733\\[-3pt]{\scriptsize(± 0.218)}} \\
    \hline

    Mistral 7B
    & \makecell{\small \textbf{0.909}\\[-3pt]{\scriptsize(± 0.111)}}
    & \makecell{\small 0.716\\[-3pt]{\scriptsize(± 0.208)}}
    & \makecell{\small \textbf{0.735}\\[-3pt]{\scriptsize(± 0.101)}}
    & \makecell{\small 0.688\\[-3pt]{\scriptsize(± 0.175)}}
    & \makecell{\small \textbf{0.932}\\[-3pt]{\scriptsize(± 0.073)}}
    & \makecell{\small 0.675\\[-3pt]{\scriptsize(± 0.197)}} \\
    \hline

    Llama 3.3 70B
    & \makecell{\small 0.842\\[-3pt]{\scriptsize(± 0.128)}}
    & \makecell{\small \textbf{0.898}\\[-3pt]{\scriptsize(± 0.147)}}
    & \makecell{\small 0.740\\[-3pt]{\scriptsize(± 0.073)}}
    & \makecell{\small \textbf{0.792}\\[-3pt]{\scriptsize(± 0.122)}}
    & \makecell{\small 0.864\\[-3pt]{\scriptsize(± 0.144)}}
    & \makecell{\small \textbf{0.880}\\[-3pt]{\scriptsize(± 0.152)}} \\
    \hline

    Qwen 3 4B
    & \makecell{\small \textbf{0.868}\\[-3pt]{\scriptsize(± 0.156)}}
    & \makecell{\small 0.663\\[-3pt]{\scriptsize(± 0.207)}}
    & \makecell{\small \textbf{0.747}\\[-3pt]{\scriptsize(± 0.103)}}
    & \makecell{\small 0.621\\[-3pt]{\scriptsize(± 0.163)}}
    & \makecell{\small \textbf{0.910}\\[-3pt]{\scriptsize(± 0.119)}}
    & \makecell{\small 0.728\\[-3pt]{\scriptsize(± 0.184)}} \\
    \hline

    \end{tabular}}
    \caption{Per-model \textit{$GCS_{env}$} results for each format. Mean shown above and standard deviation in parentheses below.}
    \label{tab:gcs_env_per_model}
\end{table*}

Table \ref{tab:gcs_env_per_model} reports per-model results for the $GCS_{env}$, which jointly accounts for structural correctness and environmental efficiency, showing that incorporating environmental efficiency systematically reduces the performance gap between baseline formats (JSON, XML, YAML) and TOON across all models. This confirms that $GCS_{env}$ effectively moderates purely correctness-driven assessments by explicitly rewarding compact and energy-efficient representations, reflecting TOON's advantage in terms of environmental efficiency. Nevertheless, the results also indicate that environmental efficiency alone is generally insufficient to fully compensate for TOON’s structural correctness limitations. In the majority of model–format comparisons, baseline formats continue to achieve higher $GCS_{env}$ values, suggesting that the structural reliability of JSON, XML, and YAML remains a dominant factor under the adopted weighting scheme ($\gamma = 0.5$). Overall, while TOON excels on sustainability-related dimensions, its reduced robustness in satisfying strict structural constraints still impacts the final evaluation. Wilcoxon signed-rank tests confirm that most baseline–TOON comparisons yield statistically significant differences in $GCS_{env}$ ($p-value < 0.05$). Importantly, the few cases where no statistically significant differences are observed provide additional insight into the interaction between model capacity, format robustness, and environmental efficiency. Specifically, no significant differences emerge for JSON and XML versus TOON in Gemma 3 27B, for XML and YAML versus TOON in GPT-oss 20B, for YAML versus TOON in Llama 3.3 70B, and for XML versus TOON in Mistral 7B. These exceptions predominantly occur either in larger models or in settings where the structural correctness gap has already been substantially reduced, suggesting that environment-aware scoring can bring TOON to parity when structural reliability is sufficiently high.

These results demonstrate that $GCS_{env}$ provides a more informative evaluation criterion than correctness-only metrics, as it captures meaningful trade-offs between reliability and sustainability. At the same time, the persistence of statistically significant differences in most comparisons indicates that environment-aware evaluation does not mask limitations in structural correctness, but rather situates them within a broader and more comprehensive performance landscape. This reinforces the suitability of $GCS_{env}$ as a benchmark metric for structured generation, enabling comparisons across formats that differ not only in syntactic robustness but also in computational and environmental footprint.

\subsection{Sensitivity Analysis on $\gamma$} \label{sec:results_gamma}

\begin{figure*}[t]
    \centering
    \begin{subfigure}[t]{0.32\textwidth}
        \centering
        \includegraphics[width=\linewidth]{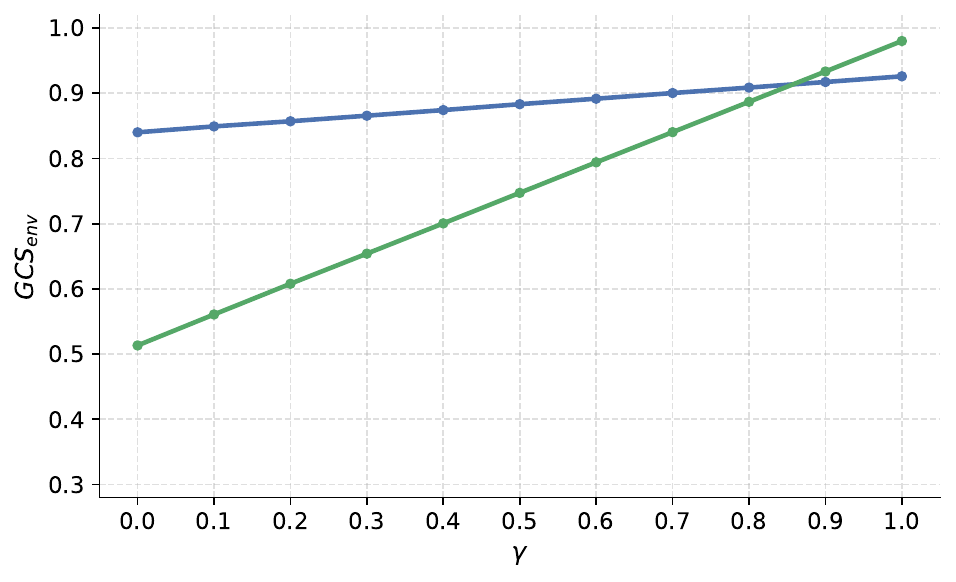}
        \label{fig:img11_gamma_sensitivity}
    \end{subfigure}\hfill
    \begin{subfigure}[t]{0.32\textwidth}
        \centering
        \includegraphics[width=\linewidth]{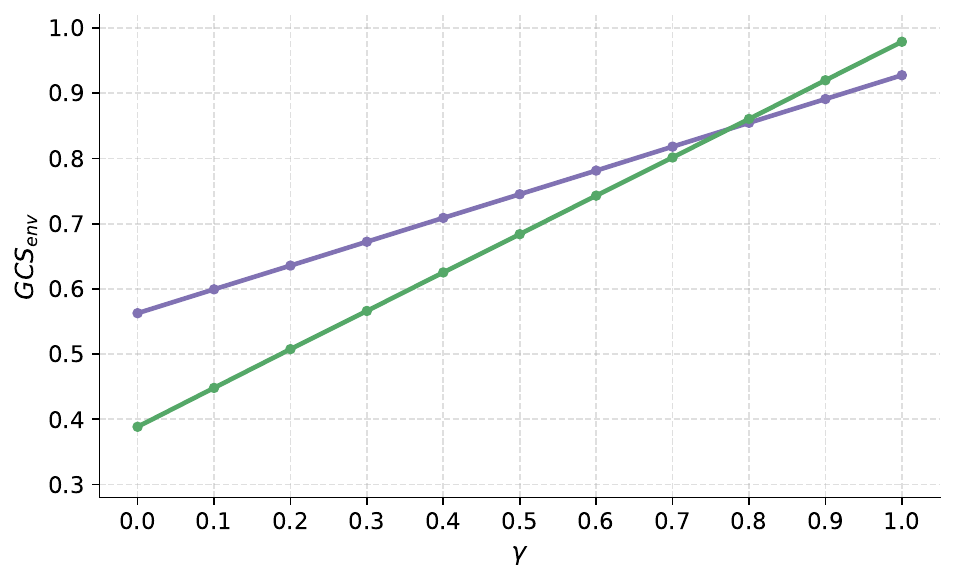}
        \label{fig:img12_gamma_sensitivity}
    \end{subfigure}\hfill
    \begin{subfigure}[t]{0.32\textwidth}
        \centering
        \includegraphics[width=\linewidth]{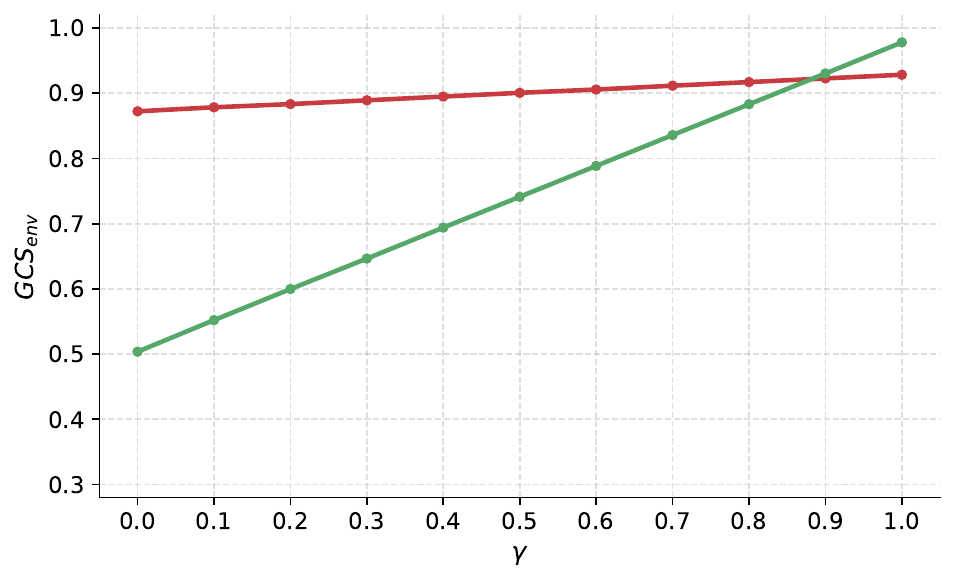}
        \label{fig:img13_gamma_sensitivity}
    \end{subfigure}
    \vspace{0.8em}
    \includegraphics[width=.4\textwidth]{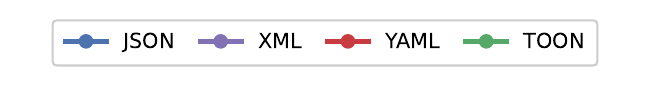}
    \caption{Sensitivity of average scores to $\gamma \in [0,1]$. Each curve corresponds to a format (JSON, XML, YAML, and TOON variants).}
    \label{fig:gamma_sensitivity}
\end{figure*}

To address \textbf{RQ3}, we analyze how the relative importance assigned to environmental efficiency influences the comparative evaluation of structured output formats. In particular, Figure \ref{fig:gamma_sensitivity} reports the sensitivity of the proposed environment-aware score $GCS_{env}$ to the weighting parameter $\gamma \in [0,1]$, which explicitly controls the relative contribution between structural correctness ($GCS$) and sustainability ($EES$). Each curve corresponds to a structured format (JSON, XML, YAML, and TOON), and scores are averaged across all evaluated models and instances.

$GCS_{env}$ exhibits a smooth and monotonic dependence on $\gamma$ for all formats, indicating that the proposed formulation behaves in a stable and predictable manner across the entire weighting spectrum. As expected, baseline formats dominate the low-$\gamma$ regime, where structural correctness is prioritized: for $\gamma \approx 0$, $GCS_{env}$ closely mirrors the correctness-only metric, and JSON, XML, and YAML consistently achieve higher scores than TOON due to their superior structural validity.

As $\gamma$ increases, the contribution of environmental efficiency becomes progressively more influential, and the relative ranking of formats changes accordingly. In particular, TOON benefits from a markedly steeper increase in $GCS_{env}$ compared to baseline formats, reflecting its consistently higher environmental efficiency. This trend highlights that TOON’s advantages are primarily driven by sustainability rather than correctness, and only emerge when environmental considerations are assigned substantial weight.

Notably, TOON surpasses the corresponding baseline formats only in the high-$\gamma$ regime, i.e., when environmental efficiency is strongly prioritized. This indicates that, under balanced evaluation settings (e.g., $\gamma = 0.5$), the gains in environmental efficiency are insufficient to fully compensate for TOON’s lower structural correctness. Conversely, in scenarios where sustainability considerations are dominant—such as large-scale deployments or carbon-constrained settings \cite{iftikhar2024reducing, liu2024green, jegham2025hungry}—TOON becomes competitive despite its reduced syntactic robustness.

Importantly, baseline formats exhibit relatively shallow slopes across $\gamma$, suggesting limited sensitivity to environmental weighting. This reflects their comparatively stable but less efficient tokenization behavior, which constrains potential gains from increasing $\gamma$. In contrast, TOON’s pronounced sensitivity to $\gamma$ underscores the fundamental role of representational compactness in driving sustainability-aware evaluation outcomes, demonstrating that the proposed $GCS_{env}$ metric enables explicit and interpretable trade-offs between correctness and sustainability.

\subsection{Summary of Findings} \label{sec:results_summary}

Across all evaluated models, tasks, and format pairs, our benchmark reveals a clear trade-off between \emph{structural correctness} and \emph{environmental efficiency}. In aggregate, TOON consistently achieves higher efficiency, generating shorter outputs (lower $N_T$) that translate into substantially lower estimated carbon emissions ($CE$) and near-ceiling environmental efficiency scores ($EES \approx 0.98$). These improvements are robust across model families and parameter scales and are statistically significant under Wilcoxon signed-rank tests. However, baseline formats (JSON, XML, YAML) remain systematically more reliable in enforcing structural constraints, yielding higher $GCS$ values, again with statistically significant differences. When combining correctness and sustainability through the proposed environment-aware score ($GCS_{env}$), the gap between TOON and baseline formats is consistently reduced, confirming that carbon-aware evaluation meaningfully moderates purely correctness-driven assessments, yet baseline formats remain superior in most settings under the adopted weighting scheme ($\gamma = 0.5$). Model-wise analyses further show that the feasibility of TOON is strongly scale-dependent: increasing model capacity progressively mitigates the lack of native format support, in some cases bringing TOON to parity with baseline formats for both \textit{GCS} and $GCS_{env}$. The sensitivity analysis on $\gamma$ demonstrates that the preferred format is not absolute but depends on evaluation priorities: TOON becomes competitive only in the high-$\gamma$ regime, where environmental efficiency is strongly prioritized.

\subsection{Discussion}

Our study highlights a consistent trade-off between structural reliability and environmental footprint in structured generation, positioning TOON as a greener representation whose practical adoption for structured generation depends on \textit{(i)} the required level of structural reliability, \textit{(ii)} the available model capacity, and \textit{(iii)} the relative importance assigned to sustainability in downstream deployment contexts. TOON yields measurable sustainability benefits such as lower emissions ($CE$) and higher environmental efficiency ($EES$), while established formats (i.e., JSON, XML, YAML) remain more robust under structural constraints, as reflected by higher \textit{GCS} values. These findings are explicitly captured by our environment-aware evaluation metric ($GCS_{env}$), which surfaces how format preferences shift as sustainability is weighted more heavily.

Importantly, these results should be interpreted in the context of realistic deployment workflows: when a generated output fails to satisfy format constraints, downstream pipelines typically require regeneration, repair, or fallback strategies. Such additional calls may reduce the environmental gains associated with more compact representations, since repeated decoding increases both token usage and emissions. In this sense, the environmental advantage of TOON is not solely a property of the representation, but also of its \emph{effective success rate} under the target output format. At the same time, our model-wise analyses indicate that sufficient model capacity can partially mitigate the lack of native support for TOON: larger models narrow, and in some cases eliminate, the correctness gap relative to baseline formats. However, this introduces a second-order effect: larger models generally exhibit higher absolute emissions, which can reduce the net sustainability benefit of adopting TOON if correctness can only be achieved by scaling up the underlying model.

From a broader perspective, our results provide evidence that representation design and model capability co-evolve: compact formats can deliver sustainability advantages, but realizing them in practice requires either stronger native support in model training or sufficient capacity to reliably satisfy previously unseen structural constraints.

\section{Conclusion}

This work investigated the trade-offs between structural correctness and environmental efficiency in structured output generation, with a particular focus on the novel, compact representation TOON. Through an extensive benchmark across multiple LLMs, parameter scales, and evaluation criteria, we addressed three complementary research questions, revealing a systematic pattern: established structured formats such as JSON, XML, and YAML remain more reliable in terms of structural validity, while TOON offers substantial efficiency and sustainability advantages due to its markedly more compact representations. These opposing properties persist across models, indicating that output format alone can have a decisive impact on both correctness and environmental footprint. Furthermore, model capacity emerges as a key moderating factor in this trade-off. Smaller and mid-sized models struggle to reliably produce structurally valid TOON outputs, whereas larger models progressively reduce (and in some cases eliminate) the correctness gap with baseline formats. This suggests that increased model capacity enables a more effective handling of previously unseen structural constraints, partially compensating for the lack of native format support. At the same time, the higher computational cost of larger models reduces the absolute environmental gains achievable through compact representations, highlighting an inherent trade-off between output correctness, efficiency, and model scale. Finally, we demonstrate that format rankings are not absolute but depend on deployment priorities and constraints, thereby confirming the need to define and employ environment-aware evaluation criteria such as our $GCS_{env}$. Such criteria are essential to enable explicit and interpretable trade-offs between structural reliability and environmental sustainability in structured generation.

\paragraph{\textbf{Limitations}}

This study has some limitations that should be considered when interpreting the results.

First, TOON generation relies on prompt-based instructions defined by the authors. Although these instructions are derived directly from the official TOON documentation, prompt design inevitably introduces a degree of subjectivity and may influence absolute performance. Alternative prompt formulations or instruction-tuning strategies could lead to different outcomes.

Second, our evaluation focuses exclusively on the \emph{generation} paradigm of the StructEval benchmark \cite{StructEval}. We do not consider \emph{conversion} tasks, where models translate between structured formats. \cite{StructEval} shows that conversion tasks are generally easier than generation tasks. As a result, our findings primarily characterize the most challenging setting for structured generation and may not fully generalize to scenarios where structured inputs are already available.

\paragraph{\textbf{Future Work}}

Several promising directions emerge from this work. First, an important extension would be to fine-tune LLMs explicitly on TOON. This would allow us to disentangle limitations due to format design from those arising purely from lack of exposure during training.

Second, future studies should investigate whether emerging foundation models begin to support new structured representations such as TOON natively through their training phase. Evaluating such models would provide insight into whether the observed trade-offs are transient or reflect more fundamental constraints in structured generation.

Finally, extending the analysis to include conversion tasks and additional structured formats would further strengthen the generality of the proposed evaluation framework and clarify how environment-aware benchmarking should be applied across different structured generation scenarios.

    
\bibliographystyle{ACM-Reference-Format}
\bibliography{bibliography}

\appendix

\end{document}